\documentclass{spie}

\usepackage{cite}

\usepackage{times} 
\usepackage{indentfirst} 

\usepackage[colorinlistoftodos,color=pink]{todonotes} 
\usepackage{xkeyval}
\presetkeys{todonotes}{inline}{}

\usepackage{graphicx}
\usepackage{hyperref}
\usepackage{color}

\usepackage{csquotes}
\usepackage{subcaption}
\usepackage{listings}
\usepackage{siunitx}
\usepackage{multirow}
\usepackage{enumerate}
\usepackage{enumitem}
\usepackage{textcomp}
\usepackage{gensymb}
\usepackage{tabularx}
\usepackage[section]{placeins}
\usepackage{xspace}
\usepackage{numprint}
\usepackage{amsmath}
\usepackage{booktabs}
\usepackage{tikz}
\usepackage{orcidlink}

\npthousandsep{\,}
\makeatletter
\DeclareRobustCommand\onedot{\futurelet\@let@token\@onedot}
\def\@onedot{\ifx\@let@token.\else.\null\fi\xspace}

\def\eg{\emph{e.g}\onedot,~} 
\def\ie{\emph{i.e}\onedot~}

\def\etal{\emph{et al}\onedot}

\newcommand{\beforemedalspace}{\hspace{0pt}}
\newcommand{\aftermedalspace}{\hspace{2pt}}
\def\medalsize{2pt}
\newcommand{\gold}{\beforemedalspace\tikz{\fill[yellow] circle(\medalsize);}\aftermedalspace}
\newcommand{\silver}{\beforemedalspace\tikz{\fill[gray] circle(\medalsize);}\aftermedalspace}
\newcommand{\bronze}{\beforemedalspace\tikz{\fill[brown] circle(\medalsize);}\aftermedalspace}
\newcommand{\nomed}{\beforemedalspace\tikz{\fill[white] circle(\medalsize);}\aftermedalspace}

\makeatother
\graphicspath{{../jpg/}{../png/}{../pdf/}{images/}}
\def\methodname{EEPPR}

\title{\methodname: Event-based Estimation of Periodic Phenomena Rate \\using Correlation in 3D}

\author{Jakub Kolář \orcidlink{0009-0005-0399-6733}, Radim Špetlík \orcidlink{0000-0002-1423-7549}, Jiří Matas \orcidlink{0000-0003-0863-4844}\\
{\normalsize kolarj55@fel.cvut.cz, spetlrad@fel.cvut.cz, matas@fel.cvut.cz}
  \skiplinehalf
  \normalsize 
  Visual Recognition Group, Faculty of Electrical Engineering, Czech Technical University in Prague, Czech Republic
  }

\begin{document}

\maketitle

\begin{abstract}
    We present a novel method for measuring the rate of periodic phenomena (\eg rotation, flicker, and vibration), by an event camera, a device asynchronously reporting brightness changes at independently operating pixels with high temporal resolution.
    The approach assumes that for a periodic phenomenon, a highly similar set of events is generated within a spatio-temporal window at a time difference corresponding to its period.
    The sets of similar events are detected by a~correlation in the spatio-temporal event stream space.
    The proposed method, \methodname, is evaluated on a dataset of $12$~sequences of periodic phenomena, \ie flashing light and vibration, and periodic motion, \eg rotation, ranging from $3.2$~Hz to 2~kHz (equivalent to $192$ -- \num{120000}~RPM).
    \methodname~significantly outperforms published methods on this dataset, achieving a mean relative error of $0.1\%$ setting new state of the art.
    The dataset and codes are publicly available on \href{https://bit.ly/EEPPR}{GitHub}.
    
\end{abstract}
\section{Introduction}
\label{sec:intro}
Accurate measurement of periodic phenomena is important in various scientific and industrial fields. Precise quantification of rotational speed, for instance, finds applications in (i) monitoring rotating machinery components for performance evaluation and quality control~\cite{motor_speedtesting}, (ii) ensuring flight stability and manoeuvrability in drones~\cite{singh_drones_2018}, (iii) analysing sports equipment like ball tracking~\cite{kamble2019ball,pingpong_cvpr2024}, and (iv) optimising energy production in wind turbines~\cite{windy_physics}.

Traditional methods for measuring periodic phenomena often involve contact-based devices such as rotary encoders and mechanical tachometers. These approaches, while established, have inherent limitations: (i) physical contact with the target object can influence its movement and introduce measurement inaccuracies, (ii) contact methods may not be feasible in scenarios with delicate objects, confined spaces, or situations where interfering with the target significantly affects its movement. Laser tachometers offer a less invasive and highly accurate\footnote{See \url{https://meters.uni-trend.com/product/ut370-series/\#Specifications}.} alternative but require reflective material (\eg a~sticker) on the target and precise aiming of the laser, as missing the reflective material pass-through results in an inaccurate measurement.
Although contactless event-based methods exist, they are either handcrafted for a specific type of periodic phenomena~\cite{EBASM_2022} or require user supervision~\cite{freqcam2022,prophesee_vibr_estim}.

In this paper, we propose a non-contact method for measuring the properties of any periodic phenomena using an event camera. 
The input of our method is the event stream, and the output is a single scalar, the rate of the periodic phenomena calculated as a robust point estimate within a spatial window.
The method operates remotely, without requiring any modification of the observed object (\eg a marker).
The approach assumes that similar sets of pixel activations of an event camera (events) will occur at regular time intervals in any sequence capturing a periodic phenomenon. 
These intervals correspond to the period. 
For phenomena where the time of repetition varies, the method provides the time of each period which can be further analysed. 

The method was evaluated on a new dataset of twelve sequences of four types of periodic phenomena: (i) flickering light, (ii) object vibration, (iii) rotation, and (iv) repetitive translational movement. A mean relative error of $0.1\%$ was achieved, which is $\approx140\times$ lower than the mean relative error of the best available published method (see Tab.~\ref{tab:combined_res}). 

\begin{figure}
\setlength{\tabcolsep}{1pt} 
    \centering
    \begin{subfigure}[b]{0.56\textwidth}
        \centering
        \includegraphics[width=\linewidth]{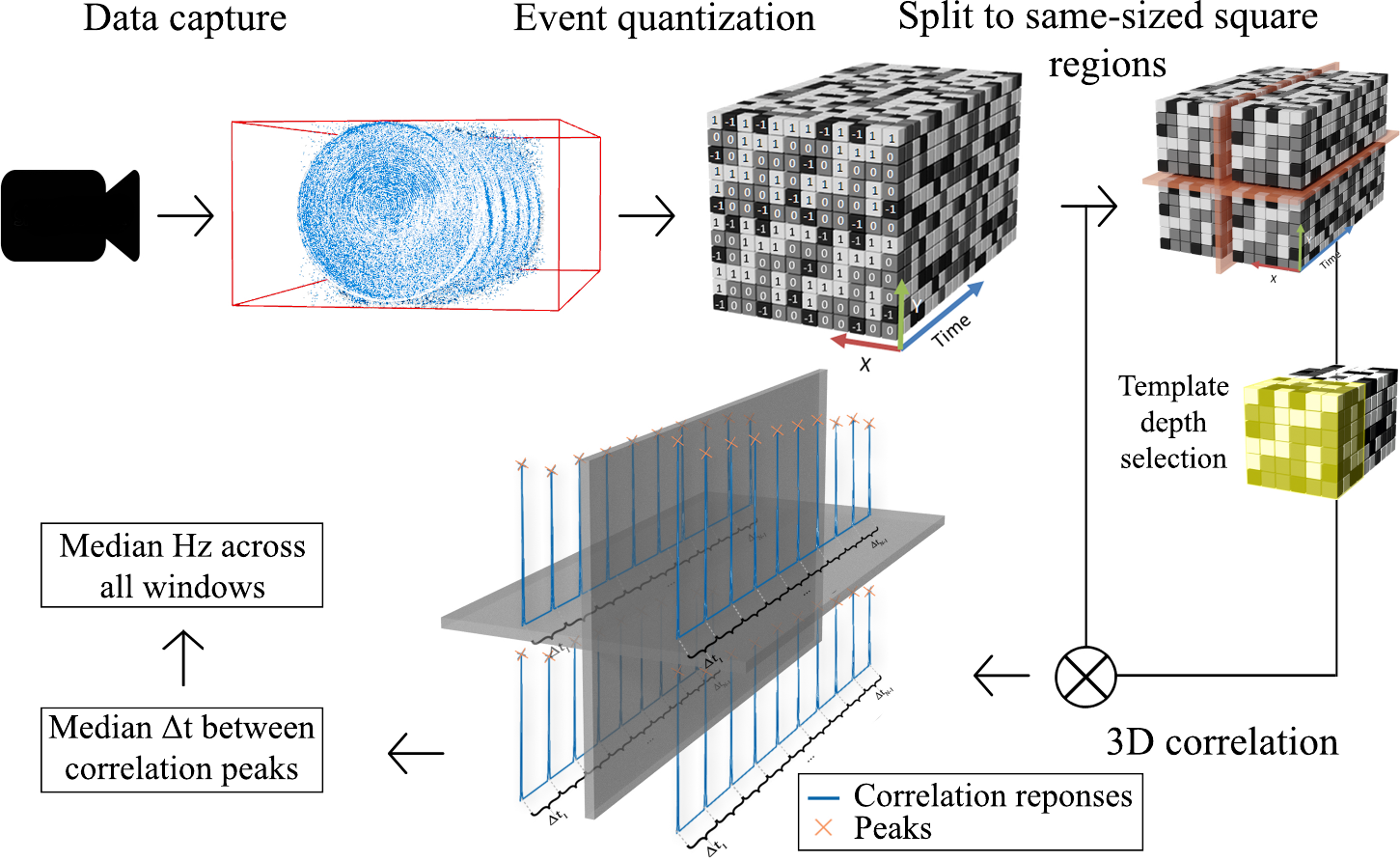}
        \caption{\methodname~method overview}
        \label{fig:diagram}
    \end{subfigure}
    \hfill
    \begin{subfigure}[b]{0.4\textwidth}
        \begin{subfigure}[b]{\linewidth}
            \centering
            \hspace{-1em}\includegraphics[width=\linewidth]{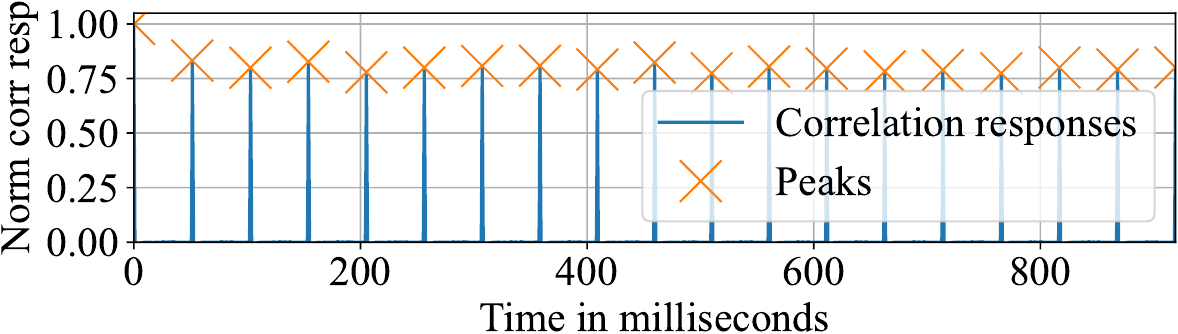}
            \vspace{-0.5em}
            \caption{Normalised correlation responses}
            \vspace{1em}
            \label{fig:large_peaks}
        \end{subfigure}
        \begin{subfigure}[b]{\linewidth}
            \centering
            \hspace{-1em}\includegraphics[width=\linewidth]{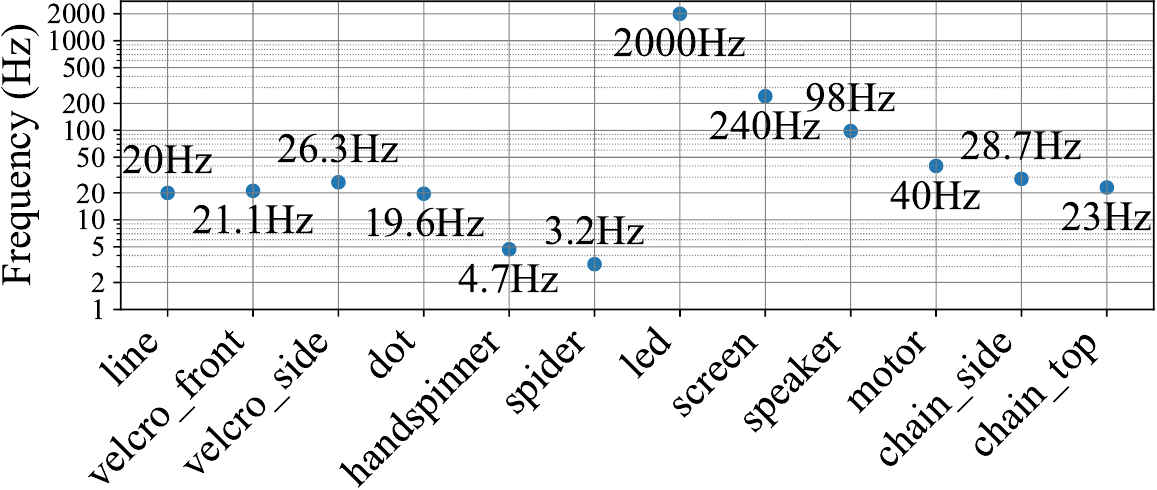}
            \vspace{-0.5em}
            \caption{Ground-truth rates}
            \label{fig:GT_freqs}
        \end{subfigure}
    \end{subfigure}
    \vspace{0.5em}
    \caption{
    (a) \methodname: 
        (i)~data captured from an event camera is aggregated into a 3D array,    
        (ii)~the array is split into same-sized areas, and in each area a template depth is automatically selected,
        (iii)~a correlation of the template with the event stream is computed in 3D,
        (iv)~a~frequency is calculated from the median of time deltas measured between correlation peaks for each window,
        (v)~the output frequency is computed as a median of measurements from all windows. 
    (b) Normalised correlation responses of a selected 3D template with $1000$ ms of spatio-temporal event stream. Periodic peaks are highly distinctive, highly regular and indicate a periodic phenomenon. 
    (c) Ground-truth rates of experiments in order of appearance in this work.}
    \label{fig:combined}
    \vspace{-1em}
\end{figure}
\vspace{-0.5em}
\section{Related work}
\label{sec:relatedwork}
In this section, we discuss existing approaches and technologies, mainly in the domain of rotation speed measurement, as this is the most commonly examined area of application.

Commercially available solutions for rotation speed measurement include contact and contactless methods. Contact-based devices include traditional mechanical tachometers that connect directly to a rotating shaft. These can introduce potential inaccuracies due to added mass and friction.
Electrostatic sensors offer a contactless approach, detecting changes in the electromagnetic field around a rotating object (equipped with a bearing). The rotational speed is the rate of these changes~\cite{electrostatic_sensor}. Optical encoder tachometers use a photoelectric sensor to detect light passing through a rotating disc with patterned segments. The rotation speed is the frequency of light intensity changes~\cite{optical_encoder}. Finally, laser tachometers utilise a laser beam that bounces off reflective markers on the rotating target. The rotation speed is the frequency of the light reflections detected by a sensor.

Several camera-based methods for rotational speed measurement have been proposed. Wang~\etal~\cite{wang_2017} proposed a~frame-by-frame 2D correlation analysis for measuring rotation speeds up to $700$~RPM with relative error of $\pm1\%$ using a low-cost RGB camera and a marker on the rotor. In~\cite{wang_2018}, multiple image similarity methods were used to analyse similarities between consecutive frames and Chirp-Z transform to calculate the rotation speed, extending the measuring range to $900$ RPM. Natali~\etal~\cite{natili_2020} further extended the measuring range by using 2D correlation and short-time Fourier Transform for measuring rotation speeds up to $1500$ RPM, highlighting the camera frame rate as the main limitation for measuring faster rotation.
While being non-contact, RGB camera-based methods are often limited by the frame rate of the camera. Additionally, markers required by~\cite{wang_2018,wang_2017} limit applicability.

The next group of methods utilise event cameras with a temporal resolution two orders of magnitude higher than RGB cameras.
The EB-ASM method~\cite{EBASM_2022} measures rotation speeds up to \num{8500}~RPM by calculating the elapsed time between spikes of distinct polarity events caused by rotating patterns in a manually selected region. However, the method struggles when noise is present, and targets lack high-contrast features. 
In EV-Tach~\cite{evtach_2022} method, only rotating objects of centrosymmetric shapes (\eg propeller blades) are assumed, and the method allows for rotation speed measurements of up to \num{6000} RPM with a relative mean absolute error as low as $0.3$\textperthousand.
The Frequency-cam~\cite{freqcam2022} method uses a second-order digital infinite impulse response (IIR) filter to reconstruct per-pixel brightness levels. The time deltas between zero-level crossing are then used to estimate the rotation speed.
In~\cite{Lv_2024}, two vibration estimation methods using high-contrast markers are proposed.
Both apply the Fast Fourier Transform (FFT) to identify the dominant frequency. The first analyses marker displacement, achieving a maximum relative error of $1.43\%$ for vibrations up to $110$ Hz, 
the second analyses event generation rate variation within a bounding box, which moves with the marker's centroid, with a maximum relative error of $1\%$ for vibrations up to $190$ Hz.
In~\cite{pingpong_cvpr2024}, a~method for estimation of a table tennis ball spin was proposed. The method uses ordinal time surfaces for tracking, the spin of the ball is estimated from optical flow of events generated by a logo on the ball. The method achieved a~spin magnitude mean error of $10.7 \pm 17.3$~RPS and a spin axis mean error of $32.9 \pm 38.2\degree$ in real time for a flying ball.
In~\cite{kolavr2024ee3p}, accurate frequency estimation of periodic phenomena lacking high-contrast features was demonstrated using a~2D correlation of spatial patterns. However, their method requires manual selection of correlation areas, whereas in our proposed method, no user intervention is necessary.

We propose two additional baseline methods for evaluating the performance of our proposed method. Those two methods are inspired by the concepts presented in the works mentioned above and are described in the following section.
Note that Prophesee, the manufacturer of event cameras, provides a method for estimating the frequency of periodic phenomena~\cite{prophesee_vibr_estim}. Although their approach is unknown, we present the results of their method in our experiments.
\begin{table}
    \centering
    \begin{minipage}{.54\linewidth}
        \centering
        \vspace{0pt}
        \caption{\methodname, average and maximum relative errors (\%) as functions of window size on the dataset \ref{sec:dataset}.}
        \resizebox{\linewidth}{!}{
        \begin{tabular}{|c||r|r|r|r|}
        \hline
            Window size (px)    & $30\times30$ & $45\times45$ & $60\times60$ & $75\times75$ \\ \hline \hline
            Avg relative error       & $37.78\hspace{0.2em}$                 & $\mathbf{19.34}$& $42.96$         & $487.98$       \\ \hline
            Max relative error       & $\mathbf{3225.00}$          & $3362.50\hspace{0.1em}$         & $7712.50$        & $22737.50$       \\ \hline
        \end{tabular}
        }
        \vspace{0.5em}
        \label{tab:window_err}
    \end{minipage}%
    \hfill
    \begin{minipage}{.42\linewidth}
        \centering
        \vspace{0pt}
        \caption{\methodname, dependence on template event count (see Sec.~\ref{sec:EEPPR-3D}) on the dataset \ref{sec:dataset}.}
        \resizebox{\linewidth}{!}{
        \begin{tabular}{|c||r|r|}
        \hline
            Template ev count   & $100-1400$ & $1500-3000$        \\ \hline \hline
            Avg relative error & $308.46$  & $\mathbf{5.75}$    \\ \hline
            Max relative error & $22737.50$  & $\mathbf{269.36}$ \\ \hline
        \end{tabular}
        }
        \vspace{0.5em}
        \label{tab:evcount_err}
    \end{minipage}\vspace{-1em}
\end{table}
\vspace{-0.5em}
\section{Baselines}
\label{sec:baselines}

\paragraph{A Simple baseline} 
was implemented to test the hypothesis that separate per-pixel events analysis is insufficient for accurate estimation of rate of complex periodic phenomena. The event camera produces a sequence of tuples $(x, y, p, t)$, where $x, y$ are spatial coordinates, $p$ is event polarity (positive~--~$1$ or negative~--~$0$), and $t$ is a timestamp in microseconds of the event creation. 
For each pixel at position $x, y$, we construct a list of timestamps $T_{x,y,p}$ of a given polarity $p$.

The baseline method then proceeds in three steps:
\begin{enumerate}[noitemsep, topsep=0pt]
 \item \textbf{Temporal analysis} -- the delta times between consecutive timestamps in $T_{x,y,p}$ are calculated, and their median is computed (\emph{temporal analysis result}). This is performed separately for each pixel and polarity, producing $2\times W^2$ results per window size $W$.
 \item \textbf{Spatial analysis} -- temporal analysis results are aggregated using the median, which produces a single \emph{spatial analysis result} per window, representing the estimated period of the phenomenon in microseconds.
 \item \textbf{Rate prediction} -- the final predicted rate for each window is derived from the estimated period $T$ in microseconds using the following equation $\nu(T) = {10^{6}}/{T}$, 
 where $\nu(T)$ returns rate in Hertz (Hz).
\end{enumerate}
\vspace{-1em}
 
\paragraph{An FFT baseline}
estimates per-pixel rate in four steps:
\begin{enumerate}[noitemsep, topsep=0pt]
    \item \textbf{FFT computation} -- for every pixel and selected polarity, we compute a one-dimensional n-point Discrete Fourier Transform~\cite{FFT_1965} with the event creation time represented by a Dirac pulse.
    \item \textbf{Signal smoothing} -- the Hanning window is applied before the Fourier transform.
    \item \textbf{Median filtering} -- in a local 3x3 window is applied to the 2D array of estimated per-pixel rates to mitigate outliers.
    \item \textbf{Results aggregation} -- the mode of detected rate is returned.
\end{enumerate}
\vspace{-0.5em}
\section{\methodname~-- The proposed method}
\label{sec:EEPPR-3D}

Our method (see Fig.~\ref{fig:diagram}) detects periodic similarities along the time axis of event data in four steps:
\begin{table}[h]
    \centering
    \resizebox{\textwidth}{!}{%
    \begin{tabular}{cccc}
        \includegraphics[height=\textheight]{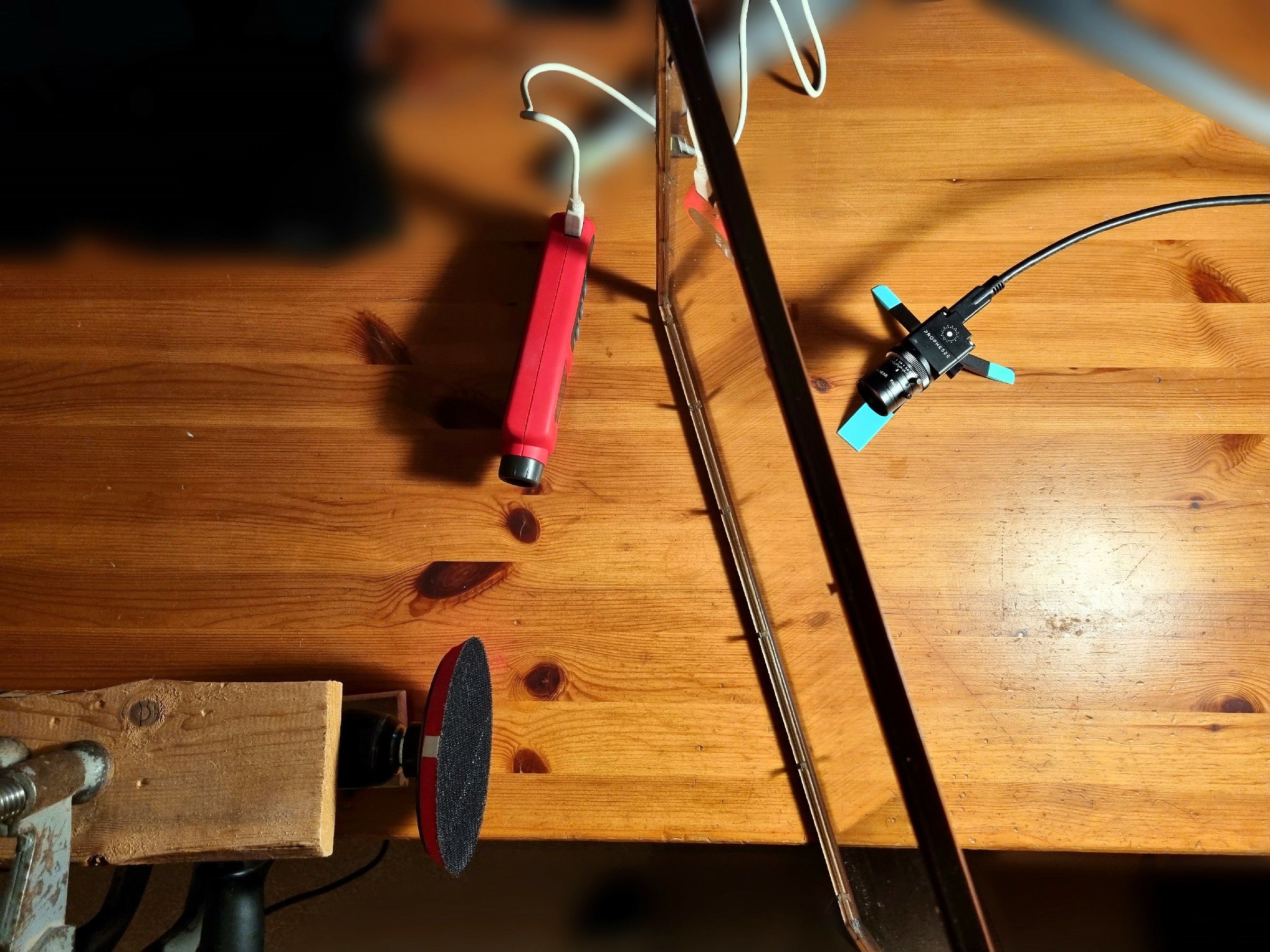} &
        \includegraphics[height=\textheight]{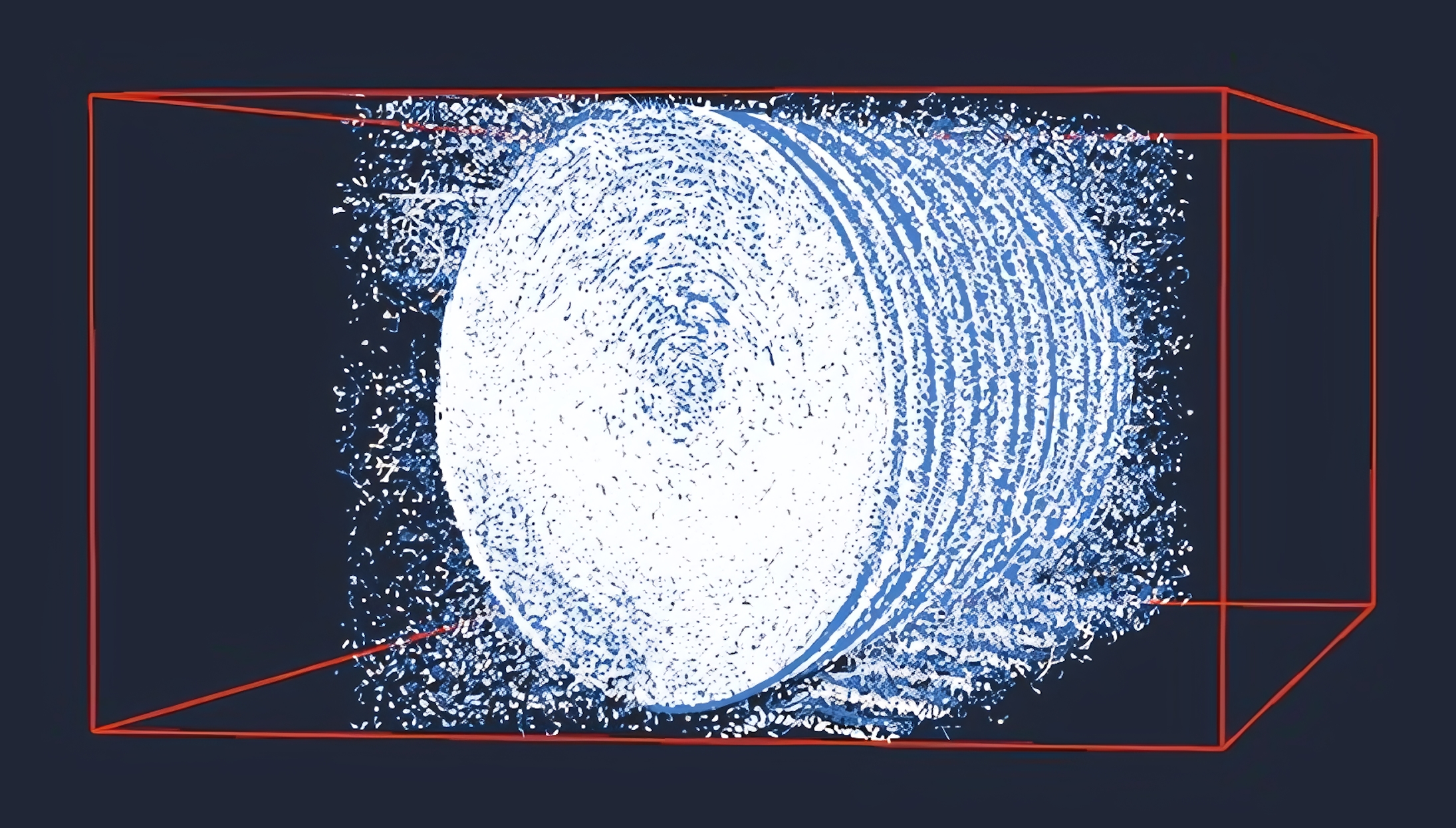} &
        \includegraphics[height=\textheight]{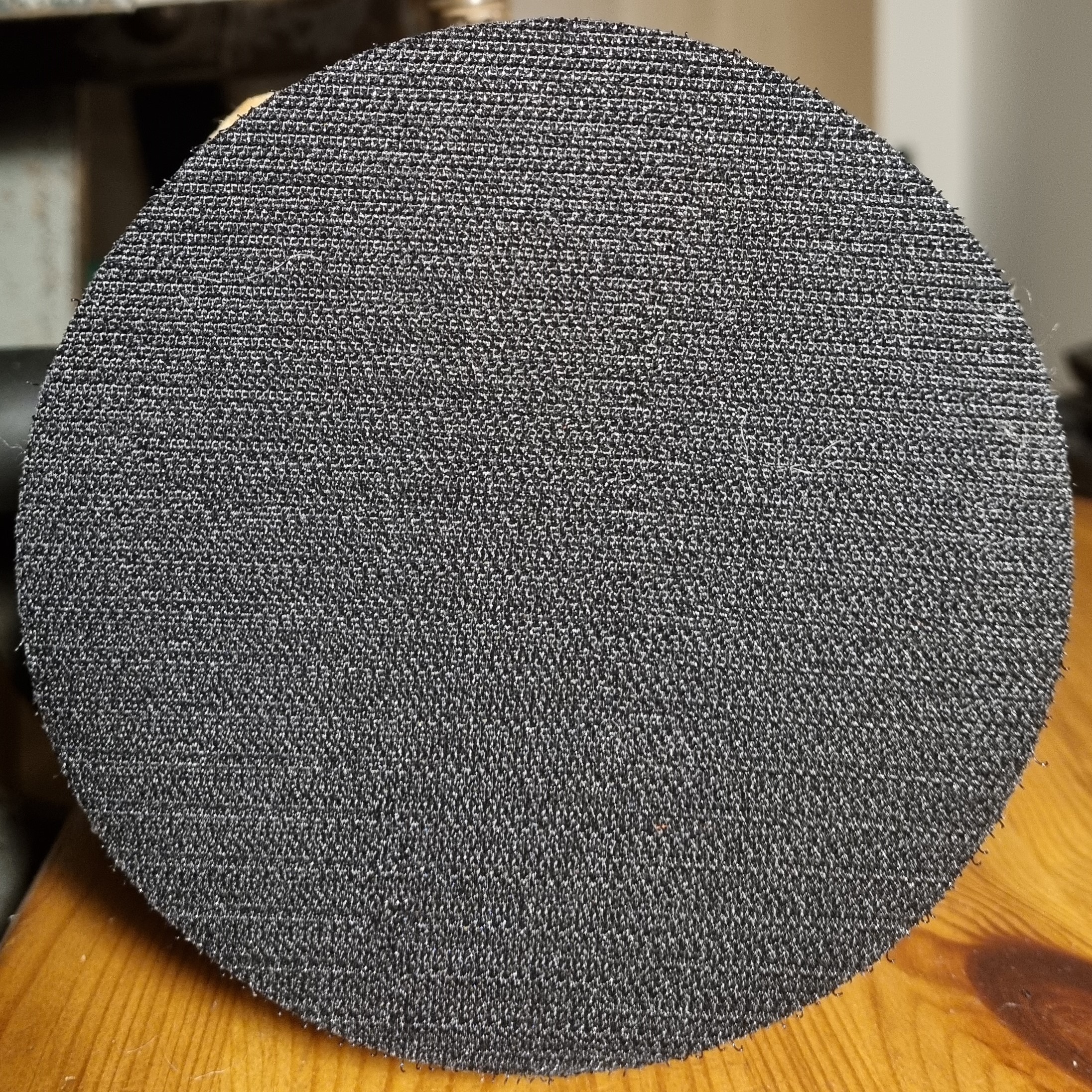} &
        \includegraphics[height=\textheight]{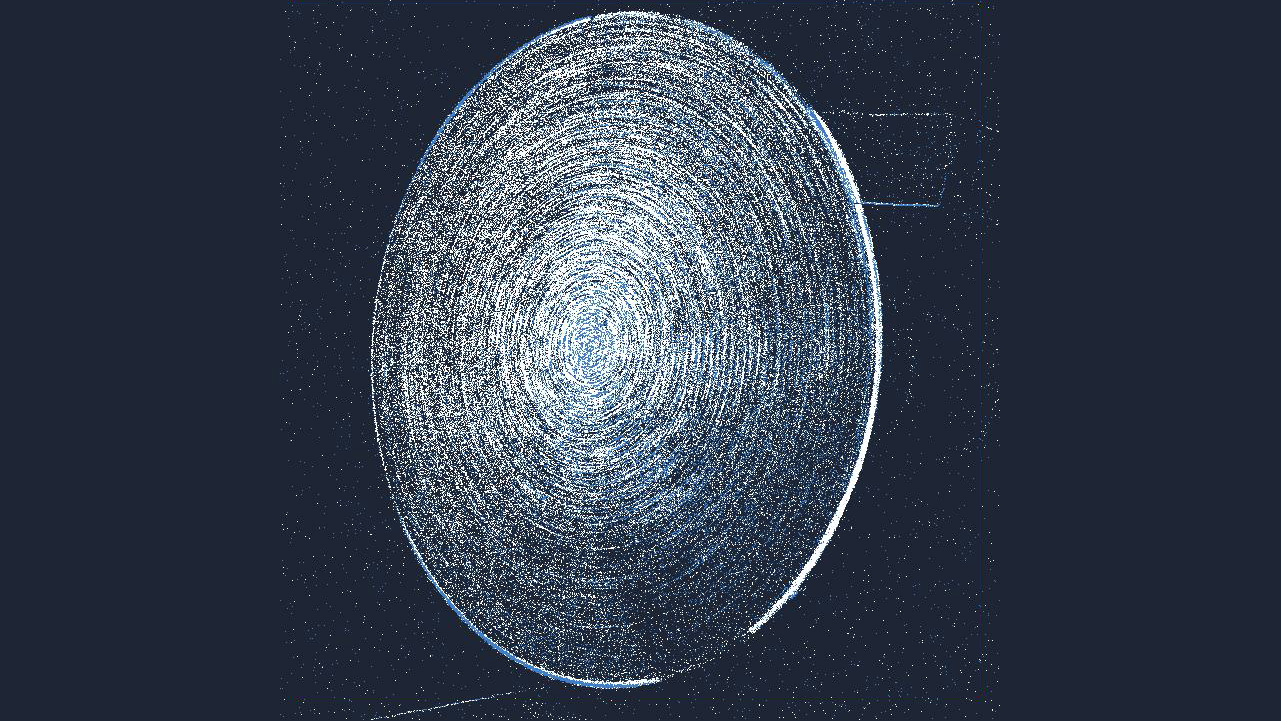} \\
    \end{tabular}%
    }\\
    \hspace{2em} (a)~Physical setup \hspace{3em} (b)~XYTime visualisation \hfill (c)~Target close-up \hspace{2.5em} (d)~Aggregated events \hspace{3em}
    \vspace{0.5em}
    \captionof{figure}{Velcro disc with a non-frontal camera behind a glass sheet (see Sec.~\ref{sec:velcro_side}): (a)~physical setup; (b)~events from a 250-millisecond window visualised in spatio-temporal space; (c)~a close-up photo of the target; (d)~aggregated events captured by the event camera ($1280\times720$px) within a time window of length equal to the period of the observed phenomenon. Positive events are represented by white colour, negative events are bright blue.}
    \label{fig:velcroside_vis}\vspace{-1em}
\end{table}

\begin{enumerate}[noitemsep, topsep=0pt]
    \item \textbf{Event quantization} -- a {3D spatio-temporal array} $\mathbf{E}$ is created by aggregating sparse events
    within non-overlapping time intervals of length $t_{\text{quant}}$ into a zero-initialized array along the third axis. For each quantised interval, the polarity ($1$ for positive, $-1$ for negative) of the last event in that interval is recorded for each pixel.
    \item \textbf{Correlation template selection} -- the array $\mathbf{E}$ is split into a grid of $K$ spatially non-overlapping areas $\mathbf{A}_i$, $i~\in~\{1, \dots K\}$, of size $W \times W$. In each area, the depth (time axis size) of the correlation template of the area $\mathbf{T}_i$ is found by thresholding on a predefined number of events $N$.
    \item \textbf{Correlation in 3D} -- a correlation in 3D of the template $\mathbf{T}_i$ 
    is computed with the area $\mathbf{A}_i$ (see Fig.~\ref{fig:large_peaks}).
    The time difference between consecutive peaks in correlation responses is the estimated period of the phenomenon. 
    \item \textbf{Results aggregation} -- the output of the method is the median of estimated periods from all areas $\mathbf{A}_i$.
\end{enumerate}
\vspace{-0.5em}
\section{Experiments}
\label{sec:experiments}

In the experiments, we used the \texttt{Prophesee EVK4 HD} event camera with $1280\times720$ resolution, capturing up to \num{1066} million events per second~\cite{ev_cam_specs} under light conditions ranging from $0.08$ up to \num{100000}~lux\footnote{Spec sheet available at \url{https://bit.ly/Sony-Prophesee-IMX636-IMX637-specsheet}.} with microsecond time resolution (equivalent to \num{10000} frames per second).

The \texttt{IMX636} sensor within the event camera allows fine-tuning its behaviour through five adjustable biases\footnote{Described in detail at \url{https://docs.prophesee.ai/stable/hw/manuals/biases.html}.}.
Contrast sensitivity thresholds, controlled by \verb|bias_diff_on| and \verb|bias_diff_off|, determine how much brighter or darker the scene needs to become before a pixel generates an event. In our experiments, we set them to $50$ and $30$, respectively, due to a higher noise of events with positive polarities. Bandwidth biases, \verb|bias_fo| and \verb|bias_hpf|, control low-pass and high-pass filters, respectively, influencing how the sensor responds to rapid or slow light changes. Finally, the dead time bias (\verb|bias_refr|) regulates the time of non-responsiveness of a pixel after generating an event. We left the remaining biases at their default settings.

Since the periodic phenomena in our dataset are stationary, and there is no ego motion of the camera, we perform 3D correlation only along the time axis -- ``correlation in 3D''. However, in cases where the actual 3D correlation would be beneficial, it can be applied without any issues.

We evaluated the accuracy of the Simple baseline method (see Sec.~\ref{sec:baselines}) when using window sizes $W$ of $30\times30$ px, $45\times45$ px, $60\times60$ px and $75\times75$ px and achieved the lowest mean relative error (MRE) by setting $W$ to $45\times45$ px. Further increasing or decreasing window size or using mode instead of the median in the temporal or spatial analysis steps of this method negatively impacted the accuracy.
The FFT baseline method (see Sec.~\ref{sec:baselines}) achieved lower mean relative error when using timestamps of events with negative polarity (MRE of $60\%$ as opposed to $67.467\%$).

We evaluated all different configurations of parameters (window sizes $W$ of $30$, $45$, $60$ and $75$ and counts of template events $N$ of $100$, $200$, \dotso, $3000$) for the proposed method (see Sec.~\ref{sec:EEPPR-3D}).
In all our experiments, the length of the quantisation window $t_{\text{quant}}$ is set to $100$ microseconds.
When using window size $W$ set to $45\times45$ px the method achieves the lowest MRE by a substantial margin and a low maximal relative error compared to results when using other window sizes (see~Tab.~\ref{tab:window_err}).
We noticed a significant improvement in accuracy when using counts of template events $N$ higher than $1500$ (see~Tab.~\ref{tab:evcount_err}).
Using 3-fold cross-validation on the introduced dataset, the best accuracy is achieved using the window size $W$ set to $45\times45$ px and the template event count $N$ set to $1800$ events. We propose these parameter as default.

We compare the results of the proposed method both with the baseline methods (see Sec.~\ref{sec:baselines}), and published methods~--~a vibration estimation method~\cite{prophesee_vibr_estim} developed by Prophesee, manufacturer of the used event camera, the open-source Frequency-cam method~\cite{freqcam2022} and EB-ASM method~\cite{EBASM_2022} which we re-implemented. As the output of Frequency-cam is in the form of colourmaps, we modified it to output a reading for each pixel instead and computed their median. Since every method produces a different amount of measurements per one second of input data, we calculate the median of measurements to obtain a single value for a comparison with other methods. We followed parameter selection guidelines provided by the authors of all examined methods \cite{prophesee_vibr_estim,freqcam2022,EBASM_2022}.
\subsection{Dataset}
\label{sec:dataset}
To capture ground truth (GT) rotation speed data, the \texttt{Uni-Trend UT372} laser tachometer\footnote{Details at \url{https://meters.uni-trend.com/product/ut370-series}.} was used. The tachometer range is $10$ to \num{99999} RPM with a relative error of $\pm0.04\%$. We averaged all tachometer readings captured in one second ($3$ to $5$) to obtain a single reference value for comparison with the results of all methods.

However, acquiring GT data this way was not feasible for all experiments. In such cases, the EE3P method~\cite{kolavr2024ee3p} was used to analyse the data multiple times with varying parameter configurations to filter potential measurement outliers and obtain a GT rate estimate as close as possible to the actual value. The rate was then verified by manually examining the event data stream.
     
Event data for three experiments in this work originate from a public dataset\footnote{Available at \url{https://docs.prophesee.ai/stable/datasets.html}.} by Prophesee. For these experiments, we used the EE3P method~\cite{kolavr2024ee3p} with manual verification to estimate the GT rate. All other sequences were captured as a part of this work and are publicly available. The final dataset contains sequences of period phenomena with rates ranging from $3.2$~Hz to 2~kHz (equivalent to $192$ -- \num{120000}~RPM), see Fig.~\ref{fig:GT_freqs}.
Due to the non-constant rates of some periodic phenomena, we restricted our analysis to one-second segments of the event stream. This selection was based on the assumption that the change in rate within such a time frame would be negligible and not significantly impact the accuracy of our measurements.

\subsection{Measuring rotation speed}
\label{sec:rotation}
    \paragraph{Felt disc with a high-contrast line}
    \label{sec:line}
        For this experiment, we used a power drill to spin a white felt disc marked with a~black line at $1200$ RPM. Event data and tachometer readings were captured simultaneously.
        All methods were expected to perform well because of the minimal noise and high contrast features present. Results confirmed this, with all methods achieving relative error lower than $0.001\%$ (see Tab.~\ref{tab:combined_res}).
\vspace{-1em}
    \paragraph{Fronto-parallel velcro disc}
    \label{sec:velcro}
        For this experiment, we used a more challenging target: a disc covered in uniform velcro material, where pattern recognition is difficult even for the human eye. A drill spun the disc at $1266$ RPM.
        Four out of six methods estimated the rate despite the lack of clear features with relative error $<0.475\%$. The EB-ASM method achieved the worst accuracy (absolute error of $6228.9$~Hz, see Tab.~\ref{tab:combined_res}).
        The Prophesee's and our proposed method achieved the highest accuracy (relative error $<0.001$\%), demonstrating robustness in this low-feature scenario.    
    \vspace{-1em}
    \paragraph{Velcro disc with a non-frontal camera behind a glass sheet}
    \label{sec:velcro_side}
        This experiment presented an even more challenging scenario. The velcro disc was spun at $1578$ RPM and captured at a $45\degree$ camera angle by the event camera through a~glass sheet (see Fig.~\ref{fig:velcroside_vis}), simulating a possible industry application when the observed object is a rotating machine part and a~transparent visor protects the camera and machine operators. 
        Interestingly, the Simple baseline method estimated the rate very well ($0.38\%$ relative error). 
        The output frequency of the FFT baseline method was roughly twice the actual rate, likely due to the presence of centrosymmetric patterns on the velcro disc. The EB-ASM method failed to produce meaningful results.
        Even with the non-aligned camera axis resulting in elliptical object trajectories on the image plane, the method by Prophesee, Frequency-cam and our proposed methods produced accurate results ($<0.001$\% relative error, see Tab.~\ref{tab:combined_res}).
        The experiment confirms that capturing data through transparent materials does not compromise accuracy.
    \vspace{-1em}
    \paragraph{High-contrast dot}
    \label{sec:dot}
        A sequence from Prophesee's dataset of an orbiting dark dot was used, captured by an older event camera model with a lower resolution and an older sensor generation, resulting in the sequence being stored using an older file type (\verb|RAW EVT2|\footnote{See \url{https://docs.prophesee.ai/stable/data/encoding_formats/index.html}.}). The Frequency-cam could not process this file encoding, hence the lack of measurements.
        The Simple baseline method did not produce any meaningful results, and the FFT baseline method estimated a frequency approximately three times higher than the GT rate. The Prophesee's method achieved an absolute error of $0.1$ Hz, while EB-ASM and our proposed method achieved accurate rate estimates ($<0.001$\% relative error, see Tab.~\ref{tab:combined_res}).
    \paragraph{Hand fidget spinner}
    \label{sec:handspinner}
        This experiment used event data from Prophesee's dataset, capturing a spinning three-blade hand fidget spinner placed on a flat surface. Because of an older file encoding, the Frequency-cam method did not provide any measurements as its dependencies do not support reading the file. 
        Most of the methods were unable to provide any meaningful results. The FFT baseline method ($0.3$ Hz overestimation) and our proposed method were the only two that achieved good performance despite the object being centrosymmetric.
        Interestingly, the method by Prophesee produced a~rate approximately four times higher than the GT rate despite the fidget spinner having only three blades. 
        Our proposed method achieved the best overall performance ($<0.001$\% relative error, see Tab.~\ref{tab:combined_res}).
    \vspace{-1.5em}
    \paragraph{Whirling Pholcus phalangioides}
    \label{sec:spider}
        The sequence used for this experiment captured the whirling behaviour for evading predators used by the Pholcus phalangioides spider. This unique defence mechanism involves the spider rapidly whirling its body while its legs remain at the silk~\cite{spider_1990}, creating a challenging yet intriguing scenario for measuring the rotation speed of the body of the spider.
        This experiment was characterised by significant noise, dim lighting conditions, and an unstable camera, which presented one of the most challenging sequences in the dataset. A second rotating pattern was visible, caused by the shadow of the spider.
        Most of the methods failed to provide accurate results. The FFT baseline method estimated a frequency four times higher than the GT rate.
        The method by Prophesee and Frequency-cam methods produced estimates that deviated by $1$~Hz and $2.1$~Hz, respectively, from the GT rate. Finally, our proposed method was the only one to achieve an accurate result with a relative error of $<0.001$\% (see Tab.~\ref{tab:combined_res}).
\subsection{Measuring frequency of periodic light flashes}
\label{sec:flashing}
    \paragraph{Flashing LED}
    \label{sec:led}
        A simple circuit with a light-emitting diode was used for this experiment. A software oscilloscope on the Raspberry Pi Pico controller~\cite{fiala_raspberry_2023} allowed precise control of the flashing frequency and duty cycle. The LED was set to flash exactly at $2000$~Hz with a $50\%$ duty cycle ($250\micro$s of turn-on time). Note that this frequency is substantially higher than those presented in other experiments in this work or related publications (see Sec.~\ref{sec:relatedwork}).
        The sequence used for this experiment was characterised by near-synchronous event generation across most pixels. 
        This experiment is the only other experiment (after the 'Felt disc with a high-contrast line', Sec.~\ref{sec:line}) in which all methods achieved near-perfect results (relative error $<0.8\%$) with the Frequency-cam performing the worst (see Tab.~\ref{tab:combined_res})
        The EB-ASM and our proposed method achieved the lowest relative error ($<0.001$\%).
\vspace{-1.5em}
    \paragraph{Refreshing screen}
    \label{sec:screen}
        For this experiment we captured the flicker of a mobile phone screen displaying a white colour at maximum brightness. A top-to-bottom pattern is visible in the event data, likely due to Pulse Width Modulation (PWM) for backlight control. The GT rate was estimated manually, using the EE3P method and confirmed by findings in a study on the display of this phone\footnote{Available at \url{https://bit.ly/notebookcheck-SamsungGalaxyS21UltraReview}.}. 
        Although the Simple baseline method failed, all other methods achieved a relative error lower than $0.8\%$.
        The limitation of the proposed method is that it requires the aggregation of events to windows of length $t_{\text{quant}}$ (see Sec.~\ref{sec:EEPPR-3D}). This restricts both the maximal measurable frequency and the theoretical limits of precision of the method. To illustrate, in this experiment, the ground truth period is $\frac{10^6}{240} = 4166,\bar{6}$ microseconds. Therefore, the closest estimate of the period ($4200$ microseconds) results in an estimated rate of $238.1$~Hz and a relative error of $0.792\%$. The Prophesee method and the FFT baseline method achieved the lowest relative errors ($<0.001$\%, see Tab.~\ref{tab:combined_res}).

\begin{table}[h]
    \centering
    \caption{Frequency estimates (Hz) and relative errors (\%) on the 12 sequence dataset. Methods: Simple baseline (Sec.~\ref{sec:baselines}), FFT baseline (Sec.~\ref{sec:baselines}), Prophesee Vibration Estimation~\cite{prophesee_vibr_estim}, EB-ASM~\cite{EBASM_2022}, Frequency-cam~\cite{freqcam2022} and \methodname~(Sec.~\ref{sec:EEPPR-3D}).}
    \setlength{\tabcolsep}{2pt}
    \resizebox{\linewidth}{!}{
    \begin{tabular}{|r|l|@{\hspace{0.5em}}r@{\hspace{0.5em}}|rrrrrr|rrrrrr|}
        \hline
         &  &  & \multicolumn{6}{c|}{Frequency Estimates (Hz)}  & \multicolumn{6}{c|}{Relative Errors (\%)} \\
         & \parbox{8em}{Method} & \parbox{2em}{GT} & \parbox{3em}{\centering \vspace{0.3em}Simple base.\vspace{0.3em}} & \parbox{3em}{\centering FFT base.} & \parbox{4em}{\centering Prophesee Vibr. Est.} & \parbox{3em}{\centering EB-ASM} & \parbox{3em}{\centering Freq-cam} & \parbox{2em}{\centering Ours} & \parbox{3em}{\centering \vspace{0.3em}Simple base.\vspace{0.3em}} & \parbox{3em}{\centering FFT base.} & \parbox{4em}{\centering Prophesee Vibr. Est.} & \parbox{3em}{\centering EB-ASM} & \parbox{3em}{\centering Freq-cam} & \parbox{2em}{\centering Ours}\\
        \hline \hline
        
        1 & \parbox{8em}{\vspace{0.2em}Felt disc with high contrast line\vspace{0.2em}} & 20.0 & \textbf{20.0} & \textbf{20.0} & \textbf{20.0} & \textbf{20.0} & \textbf{20.0} & \textbf{20.0} & $<\epsilon$ \gold & $<\epsilon$ \gold & $<\epsilon$ \gold & $<\epsilon$ \gold & $<\epsilon$ \gold & $<\epsilon$ \gold \\
        \hline
        2 & Velcro disc (front) & 21.1 & 24.2 & 21.0 & \textbf{21.1} & 6250.0 & 21.2 & \textbf{21.1} & 14.6 \bronze & 0.5 \silver & $<\epsilon$ \gold & 29520.9 \nomed & 0.5 \silver & $<\epsilon$ \gold \\
        \hline
        3 & Velcro disc (side) & 26.3 & 26.4 & 53.0 & \textbf{26.3} & 2439.0 & \textbf{26.3} & \textbf{26.3} & 0.4 \silver & 101.5 \bronze & $<\epsilon$ \gold & 9173.8 \nomed & $<\epsilon$ \gold & $<\epsilon$ \gold \\
        \hline
        4 & High contrast dot & 19.6 & 7462.7 & 59.0 & 19.5 & \textbf{19.6} & N/A & \textbf{19.6} & 37975.0 \nomed & 201.0 \bronze & 0.5 \silver & $<\epsilon$ \gold & N/A \nomed & $<\epsilon$ \gold \\
        \hline
        5 & Fidget spinner & 4.7 & 451.2 & 5.0 & 19.1 & 1111.1 & N/A & \textbf{4.7} & 9500.0 \nomed & 6.3 \silver & 306.3 \bronze & 23540.4 \nomed & N/A \nomed & $<\epsilon$ \gold \\
        \hline
        6 & Whirling spider & 3.2 & 35.1 & 13.0 & 4.2 & 298.6 & 5.3 & \textbf{3.2} & 996.9 \nomed & 306.3 \nomed & 31.3 \silver & 9231.3 \nomed & 65.6 \bronze & $<\epsilon$ \gold \\
        \hline
        7 & Flashing LED & 2000.0 & 1996.0 & 1999.4 & 2000.1 & \textbf{2000.0} & 1984.1 & \textbf{2000.0} & 0.2 \nomed & 0.03 \bronze & 0.005 \silver & $<\epsilon$ \gold & 0.7 \nomed & $<\epsilon$ \gold \\
        \hline
        8 & \parbox{8em}{\vspace{0.2em}Refreshing screen\vspace{0.2em}} & 240.0 & 2770.1 & \textbf{240.0} & \textbf{240.0} & 239.8 & 239.9 & 238.1 & 1054.2 \nomed & $<\epsilon$ \gold & $<\epsilon$ \gold & 0.08 \bronze & 0.04 \silver & 0.8 \nomed \\
        \hline
        9 & Speaker diaphragm & 98.0 & 21.0 & 98.1 & 97.8 & 99.0 & 95.3 & \textbf{98.0} & 78.6 \nomed & 0.1 \silver & 0.2 \bronze & 1.0 \nomed & 2.8 \nomed & $<\epsilon$ \gold \\
        \hline
        10 & Vibrating motor & 40.0 & 52.2 & \textbf{40.0} & \textbf{40.0} & 370.4 & N/A & \textbf{40.0} & 30.5 \bronze & $<\epsilon$ \gold & $<\epsilon$ \gold & 826.0 \nomed & N/A \nomed & $<\epsilon$ \gold \\
        \hline
        11 & Bike chain (side) & 28.7 & 123.8 & 1.0 & 33.9 & 1111.1 & 44.7 & \textbf{28.7} & 331.4 \nomed & 96.5 \nomed & 18.1 \silver & 3771.4 \nomed & 55.8 \bronze & $<\epsilon$ \gold \\
        \hline
        12 & Bike chain (top) & 23.0 & 51.6 & 22.0 & 30.8 & 1428.6 & 22.7 & \textbf{23.1} & 124.3 \nomed & 4.3 \bronze & 33.9 \nomed & 6111.3 \nomed & 1.3 \silver & 0.4 \gold \\
        \hline
        \hline
        & Mean relative error & -- \hspace{0.2em} & -- \hspace{0.2em} & -- \hspace{0.2em} & -- \hspace{0.2em} & -- \hspace{0.2em} & -- \hspace{0.2em} & -- \hspace{0.2em} &  $4175.5$ \nomed  &  $59.7$ \nomed    &  $32.5$ \bronze   &  $6848.0$ \nomed  & $14.0$   \silver  &  $0.1$ \gold \\
        \hline
        & Max relative error & -- \hspace{0.2em} & -- \hspace{0.2em} & -- \hspace{0.2em} & -- \hspace{0.2em} & -- \hspace{0.2em} & -- \hspace{0.2em} & -- \hspace{0.2em} &  $37975.0$ \nomed &  $306.3$ \bronze  &  $306.3$ \nomed   &  $29520.9$ \nomed & $65.6$   \silver  &  $0.8$ \gold \\
        \hline
    \end{tabular}
    }
    \label{tab:combined_res}
    \vspace{-1em}
\end{table}
\subsection{Measuring vibration frequency of a speaker and a motor}
\label{sec:vibration}
    \paragraph{Speaker diaphragm}
    \label{sec:speaker}
        In this experiment, we captured and analysed vibrating diaphragm of a~speaker with two large low-frequency drivers. An Android application\footnote{Application available at \url{https://bit.ly/Frequency-Sound-Generator-APK}.} precisely controlled the emitted sound (musical note $G_{2}$, $98$~Hz). The event camera captured the vibrating diaphragm of a speaker at approximately 30\degree~angle.
        The most prominent features in the event stream originate from the vibrating edges of the diaphragm and the manufacturer's logo located in its centre.
        The proposed method was the only method that achieved a relative error lower than $0.1\%$. The FFT baseline and Prophesee methods achieved relative error under $0.2\%$, followed by EB-ASM ($1.0\%$) and Frequency-cam ($2.8\%$). The Simple baseline method deviated the most from the reference frequency ($78.6\%$ relative error); see Tab.~\ref{tab:combined_res}.
    \vspace{-1.5em}
    \paragraph{Vibrating motor}
    \label{sec:motor}
        This experiment analysed a sequence from Prophesee's dataset capturing a vibrating motor.
        The most reliable visual features were the horizontal cooling fins of the motor, which periodically moved in a vertical direction.
        The EB-ASM and the Simple baseline method deviated significantly from the GT frequency (relative error $>30\%$, see Tab.~\ref{tab:combined_res}). Because of an older file encoding, the Frequency-cam method did not provide any measurements.
        The FFT baseline, Prophesee's and our proposed method achieved a relative error lower than $0.001$\%. 

\vspace{-0.5em}
\subsection{Measuring movement frequency}
\label{sec:movement}
    \paragraph{Bike chain from side view}
    \label{sec:chain_side}
        The sequence used for this experiment captured the bicycle chain from a close-up side view. 
        The FFT baseline method correctly estimated the frequency in pixels that capture the upper and lower chain link sections, where events are periodically generated per chain link passage. However, the estimated frequency in most pixels was incorrect ($1$~Hz, see Tab.~\ref{tab:combined_res}).
        The Prophesee method deviated by $\approx18\%$ from the ground truth, being the second most accurate;
        this is one of only four experiments where the method by Prophesse had a relative error greater than $0.21\%$.
        \methodname~achieved the best performance, being the only method to produce meaningful results.
\vspace{-1em}
    \paragraph{Bike chain from top view}
    \label{sec:chain_top}
        In the final experiment, we captured the same bike chain from a top view.
        The Frequency-cam method achieved the second-best result with a $1.3\%$ relative error, with the proposed method achieving the lowest absolute error of $0.1$~Hz. The Simple baseline, Prophesee's, and EB-ASM methods showed significant deviations from the GT~frequency (relative errors $34\%$ -- $6111\%$).

    \subsection{Discussion}
    \label{sec:discussion}
        We analysed twelve sequences capturing various periodic phenomena using six different methods. The experiments where the target lacked high-contrast features were shown to be the most challenging.
        The worst performing methods were the Simple baseline method and the EB-ASM method~\cite{EBASM_2022} achieving MRE of $4175.5\%$ and $6848.0\%$ respectively (see~Tab.~\ref{tab:combined_res}). The third lowest mean relative error across all experiments ($32.5\%$) was achieved by the Vibration Estimation method by Prophesee. The Frequency-cam method achieved the second lowest mean relative error (MRE) of $14.1\%$, which is significantly higher than the MRE achieved by the proposed method ($0.1\%$).
        Note that the Frequency-cam method could not analyse three of the twelve sequences we used in this work, which originated from the public dataset by Prophesee due to an older file encoding type. All other methods were able to analyse all sequences.

\vspace{-0.5em}
\section{Conclusion}
\label{sec:conclusion}

A novel contactless event-based method, \methodname, for measuring the rate of various periodic phenomena, such as rotational speed, vibration, flicker and periodic movement, was presented.
The method makes only a single assumption that the observed object periodically returns to a known state or position, producing a similar set of events in the event stream. 
The~rate of periodic phenomena is computed by measuring time deltas between peaks detected in correlation responses of an automatically selected template correlated with a spatio-temporal event stream.
\methodname~was evaluated on a new proposed dataset of twelve sequences
that represent a wide range of periodic phenomena (light flickering, object vibration, rotational speed, and periodic movement), including targets
lacking prominent features, 
being behind transparent materials,
or having centrosymmetric shapes, 
achieving the mean relative error of $0.1\%$ setting new state of the art.
The method works in non-frontal camera placement,
measuring frequencies as high as $2000$ Hz (equivalent to \num{120000} RPM)
even when a significant amount of noise is present.
The dataset and code are publicly available on \href{https://bit.ly/EEPPR}{GitHub}.

\paragraph{Limitations} The method assumes the periodic phenomenon undergoes only periodic movement without additional motion. Handling of tracking and the template update is future work. Aggregating events within fixed intervals $t_{\text{quant}}$ restricts both the maximal measurable frequency and the theoretical limits of precision. Reducing the aggregation length $t_{\text{quant}}$ mitigates this at a cost of higher computation time. Future work will explore sparse data correlation to avoid aggregation.

\vspace{-1.0em}
{
\paragraph{Acknowledgements}
The research reported in this paper has been funded by BMK, BMAW, and the State of Upper Austria within the SCCH competence center INTEGRATE (grant no.\ 892418) part of the FFG COMET Competence Excellent Technologies Programme, and by the Czech Technical University in Prague grant No. SGS23/173/OHK3/3T/13.
}

\vspace{-0.5em}
\bibliographystyle{spiebib}
\bibliography{bibliography}

\end{document}


\maketitle
\renewcommand\thefigure{\thesection.\arabic{figure}}

\section{Experimental Setup}

For the 'Felt disc with a high-contrast line' experiment (see Sec.~\ref{sec:line}), we captured a power drill spinning a white felt disc marked with a black line, see Fig.~\ref{fig:line_vis}.

\begin{table}
\centering
\resizebox{\textwidth}{!}{%
\begin{tabular}{cc}
    \includegraphics[height=\textheight]{images/line_phys~1} &
    \includegraphics[height=\textheight]{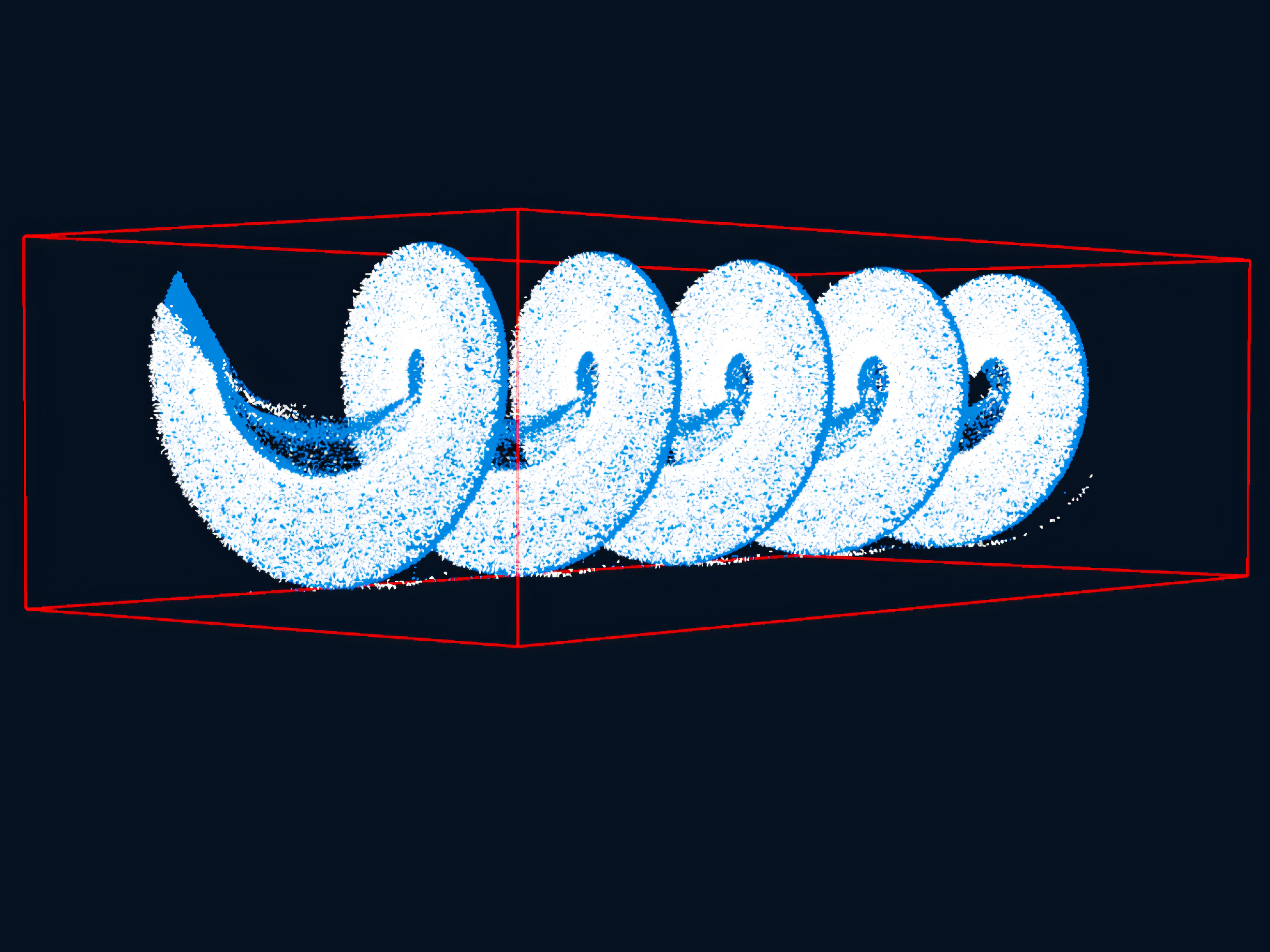}\\
\end{tabular}%
}\\
\hspace{4.5em} (a)~Physical setup \hfill (b)~XYTime visualisation \hspace{3em}

\resizebox{\textwidth}{!}{%
\begin{tabular}{cc}
    \includegraphics[height=\textheight]{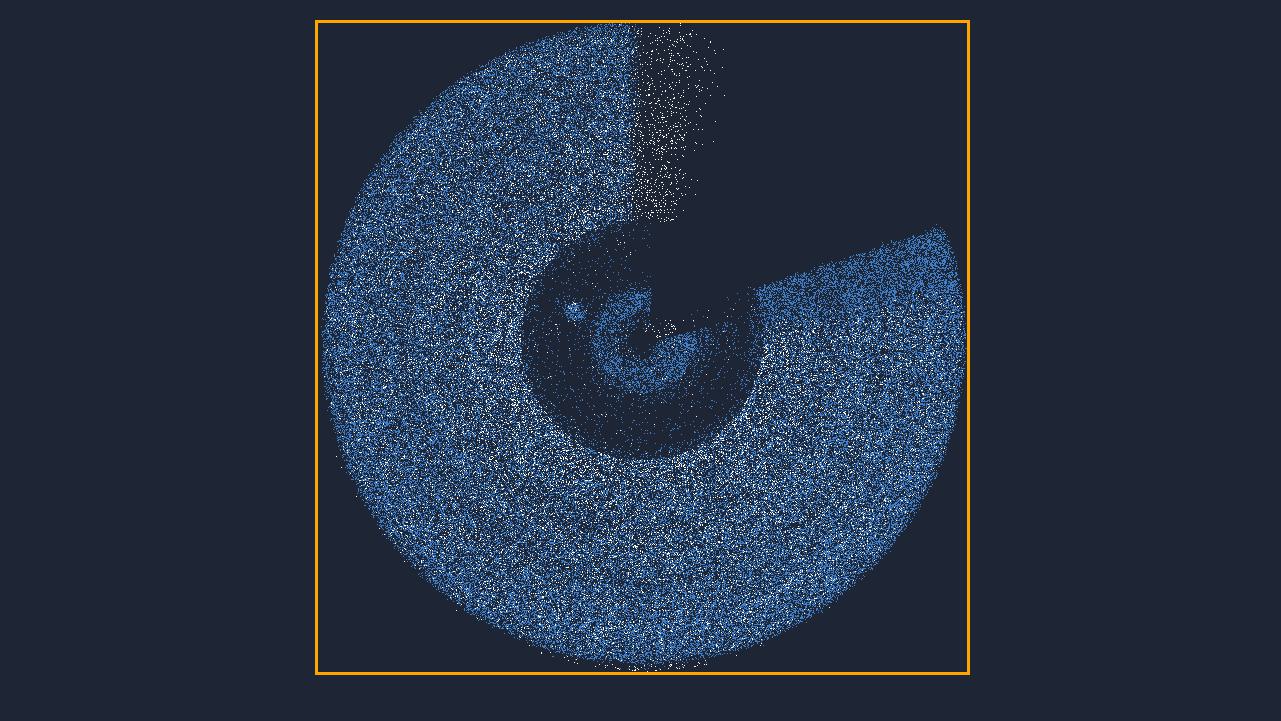} &
    \includegraphics[height=\textheight]{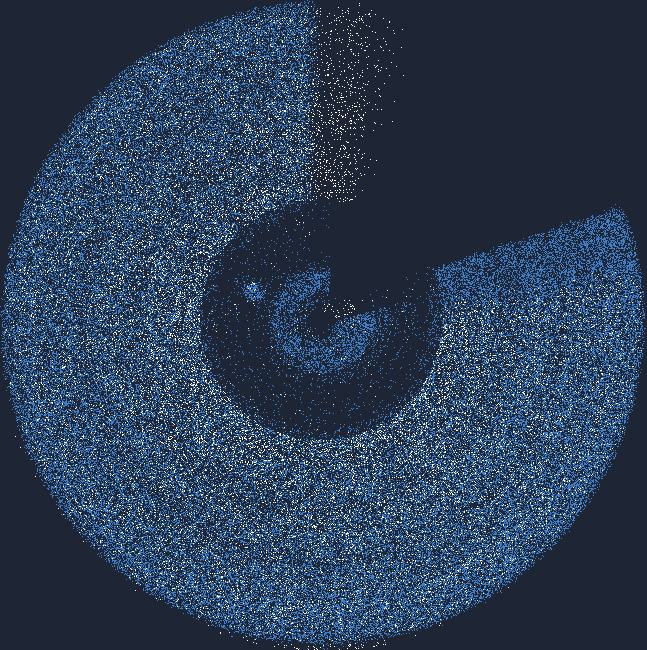}\\
\end{tabular}%
}\\
\hspace{6.5em} (c)~Aggregated events \hfill (d)~Region of Interest \hspace{1.5em}

\vspace{0.8em}\captionof{figure}{Felt disc with a high-contrast line (see Sec.~\ref{sec:line}): (a)~physical setup; (b)~events from a 250-millisecond window visualised in spatio-temporal space; (c)~aggregated events captured by the event camera ($1280\times720$px) within a time window of length equal to $\frac{8}{10}$ of the period of the observed phenomenon, highlighted region of interest ($655\times655$px) shown in (d). Positive events are represented by white colour, and negative events are bright blue.}
\label{fig:line_vis}
\end{table}
\FloatBarrier

For the 'Fronto-parallel velcro disc' experiment (see Sec.~\ref{sec:velcro}), a drill spun the disc at $1266$ RPM with the event camera facing it directly, while a laser tachometer measured speed from the side of the disc where a reflective sticker was attached (Fig.~\ref{fig:velcro_vis}a).

\begin{table}[!htb]
\centering
\resizebox{\textwidth}{!}{%
\begin{tabular}{cc}
    \includegraphics[height=\textheight]{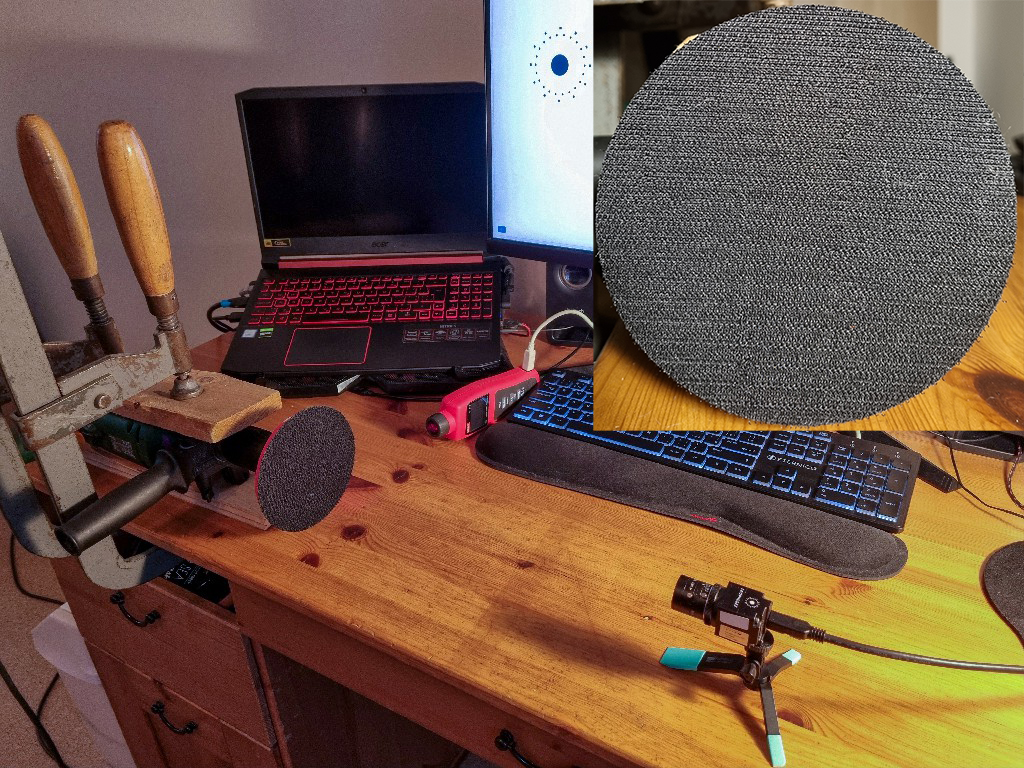} &
    \includegraphics[height=\textheight]{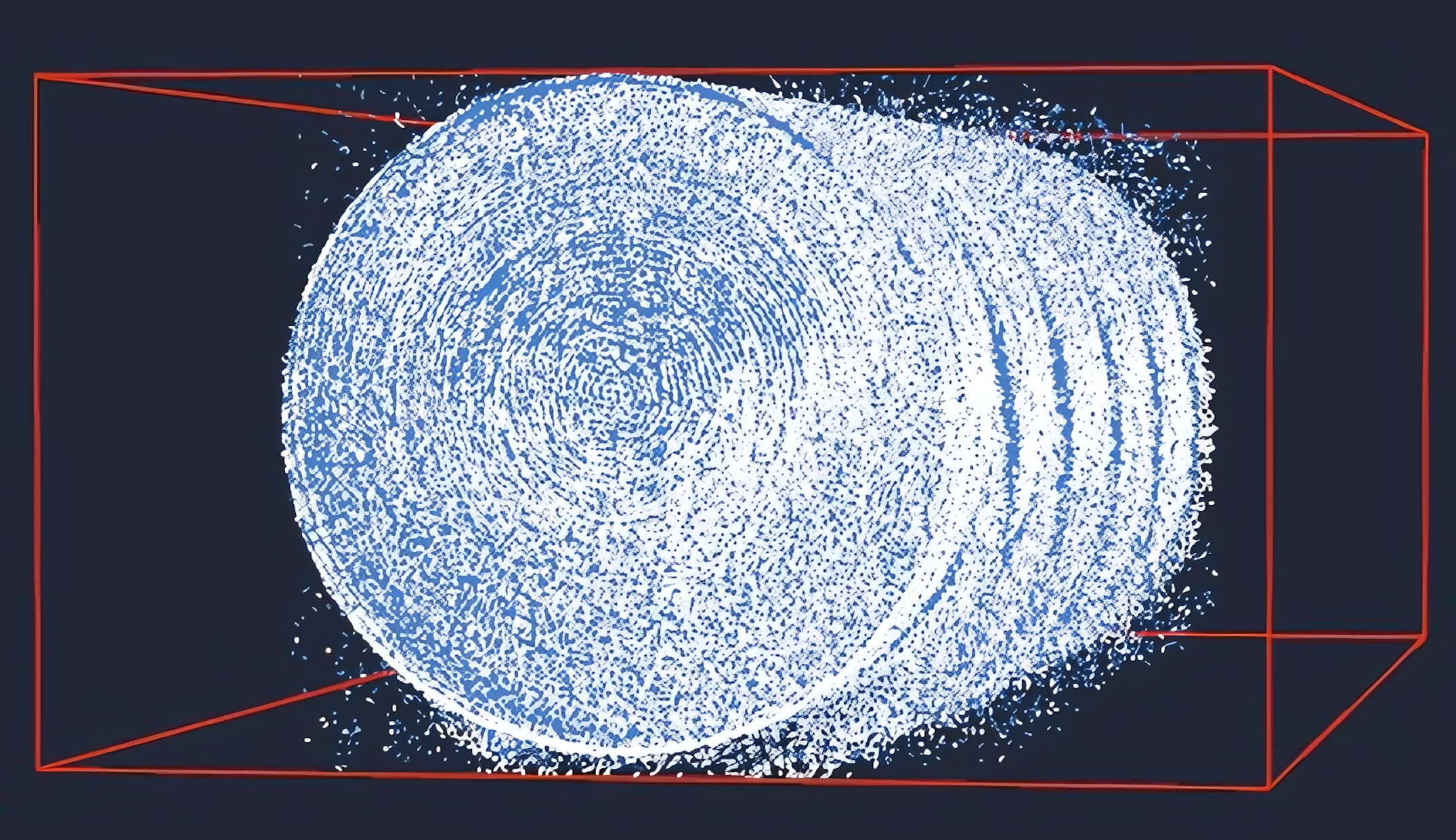}\\
\end{tabular}%
}\\
\hspace{4em} (a)~Physical setup \hfill (b)~XYTime visualisation \hspace{5em}

\resizebox{\textwidth}{!}{%
\begin{tabular}{cc}
    \includegraphics[height=\textheight]{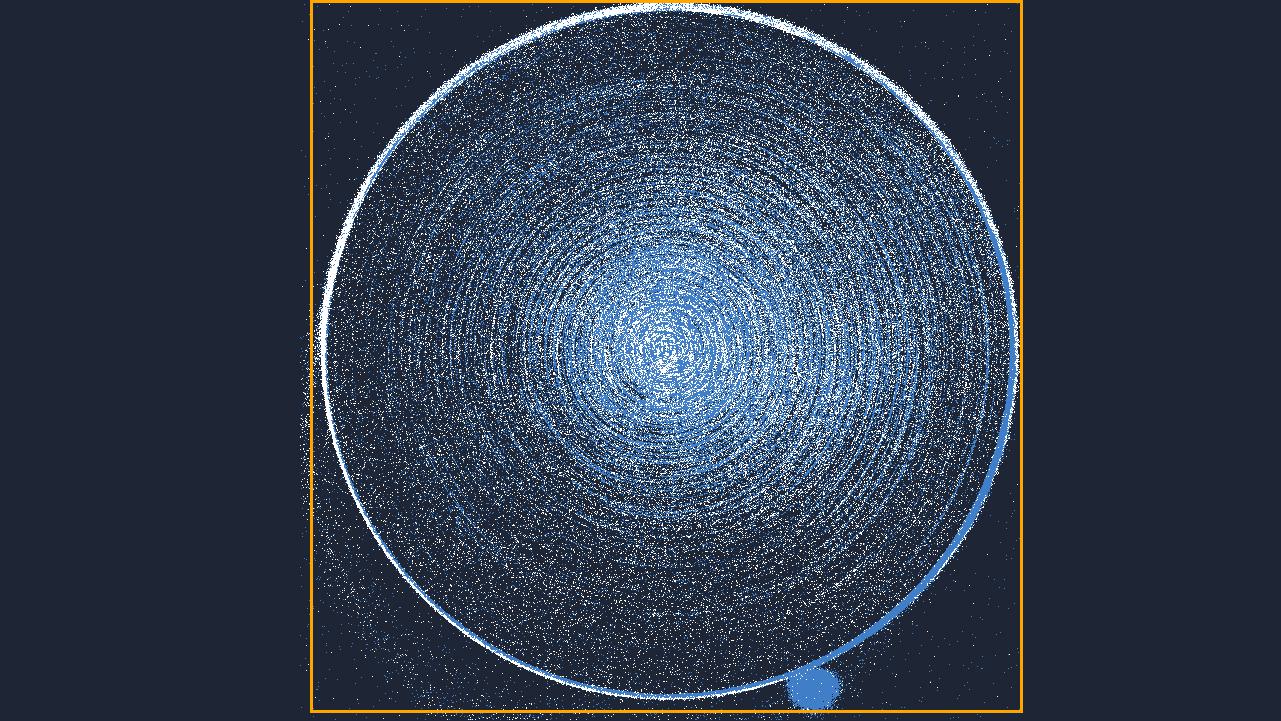} &
    \includegraphics[height=\textheight]{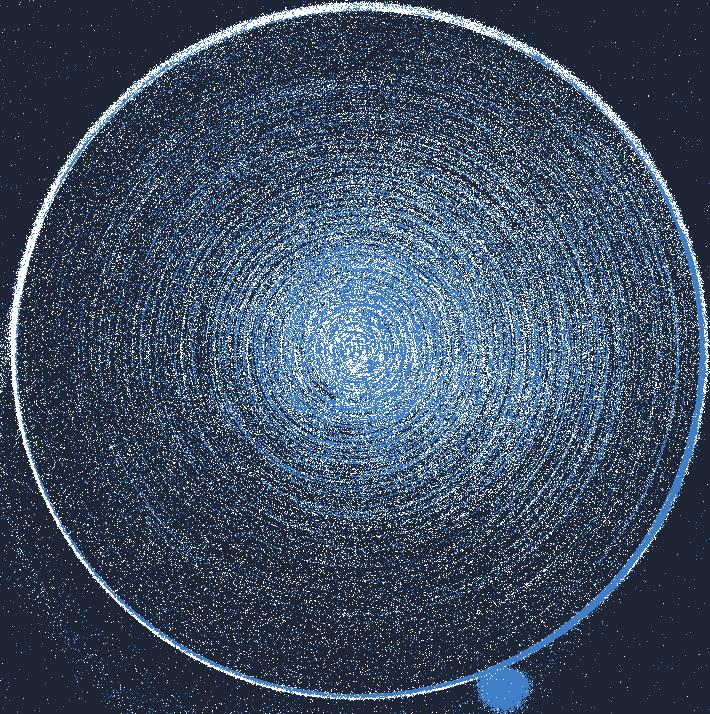} \\
\end{tabular}%
}\\
\hspace{7em} (c)~Aggregated events \hfill (d)~RoI \hspace{5em}
\vspace{0.8em}\captionof{figure}{Fronto-parallel velcro disc (see Sec.~\ref{sec:velcro}): (a)~physical setup and target detail; (b)~events from a 250-millisecond window visualised in spatio-temporal space; (c)~aggregated events captured by the event camera ($1280\times720$px) within a time window of length equal to the period of the observed phenomenon, highlighted region of interest ($713\times713$px) shown in (d). Positive events are represented by white colour, and negative events are bright blue.}
\label{fig:velcro_vis}
\end{table}
\FloatBarrier
\newpage
In the next experiment (see Sec.~\ref{sec:velcro_side}), the velcro disc was captured by the event camera through a glass sheet at $45\degree$ camera angle (see~Fig.~\ref{fig:velcroside_vis}a).

\begin{table}[!htb]
\centering
\resizebox{\textwidth}{!}{%
\begin{tabular}{cc}
    \includegraphics[height=\textheight]{images/setup_experiment3~1} &
    \includegraphics[height=\textheight]{images/velcroside_xyt2lq_upscayl_5x_remacri}\\
\end{tabular}%
}\\
\hspace{4em} (a)~Physical setup \hfill (b)~XYTime visualisation \hspace{5em}

\resizebox{\textwidth}{!}{%
\begin{tabular}{ccc}
    \includegraphics[height=\textheight]{images/velcro_detail_front.jpg} &
    \includegraphics[height=\textheight]{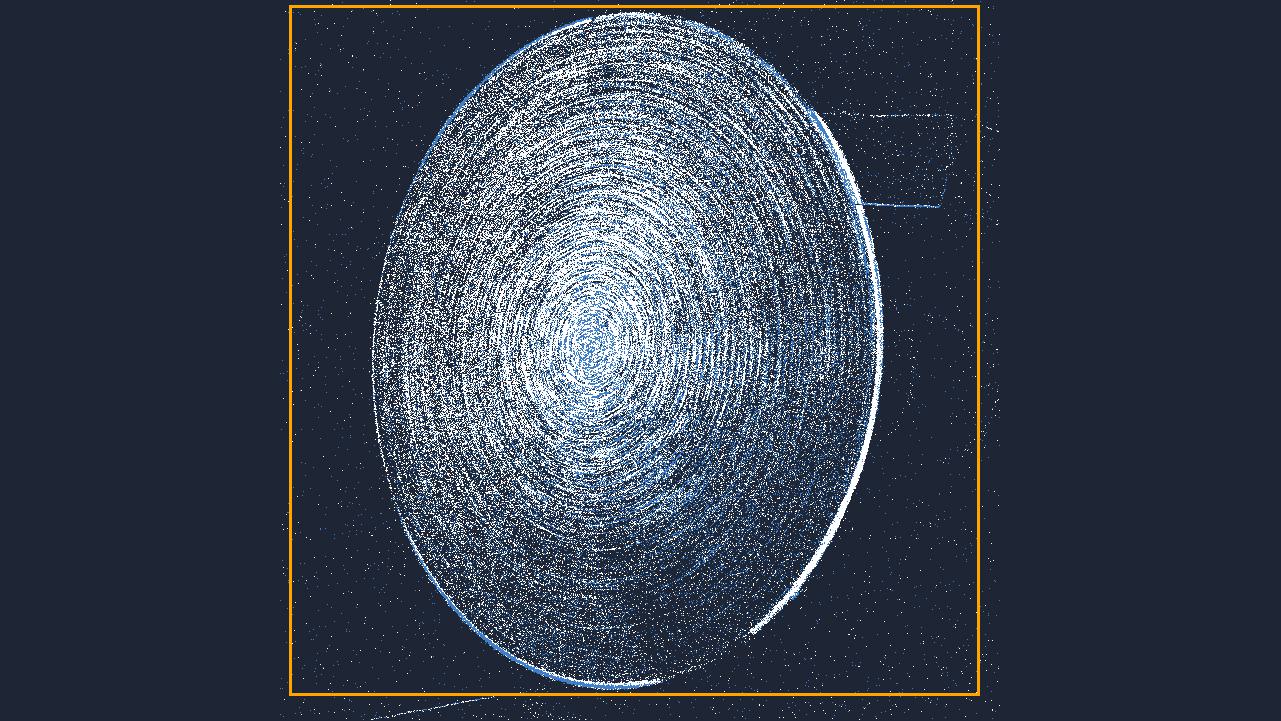} &
    \includegraphics[height=\textheight]{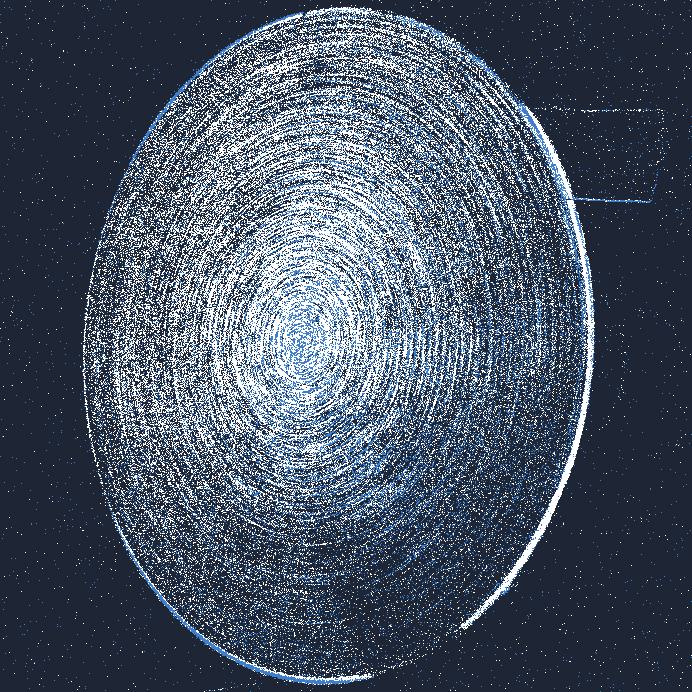}\\
\end{tabular}%
}\\
\hspace{0.5em} (c)~Target close-up \hfill (d)~Aggregated events \hfill (e) Region of Interest \hspace{0.2em}
\vspace{0.8em}\captionof{figure}{Velcro disc with a non-frontal camera behind a glass sheet (see Sec.~\ref{sec:velcro_side}): (a)~physical setup; (b)~events from a 250-millisecond window visualised in spatio-temporal space; (c)~a close-up photo of the target; (d)~aggregated events captured by the event camera ($1280\times720$px) within a time window of length equal to the period of the observed phenomenon, highlighted region of interest ($691\times691$px) shown in (e). Positive events are represented by white colour, and negative events are bright blue.}
\label{fig:velcroside_vis}
\end{table}
\FloatBarrier
\newpage
For this experiment, a sequence from Prophesee's dataset was used, capturing an orbiting dark dot (see Sec.~\ref{sec:dot}). The reference photo was generated using Prophesee's Event to Video method (see Fig.~\ref{fig:dot_vis}a).
\begin{table}
\centering
\resizebox{\textwidth}{!}{%
\begin{tabular}{cc}
    \includegraphics[height=\textheight]{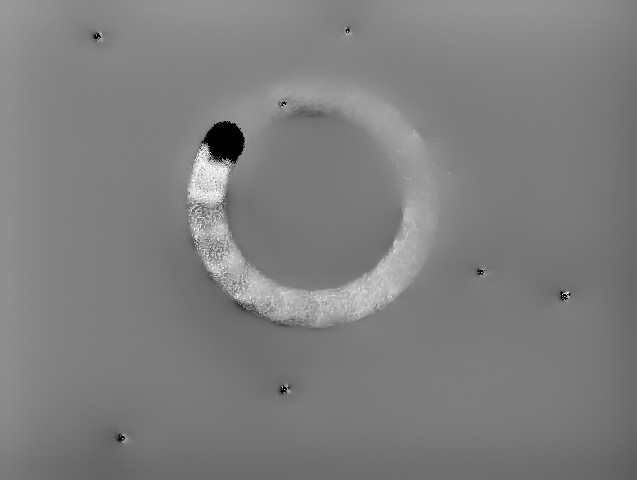} &
    \includegraphics[height=\textheight]{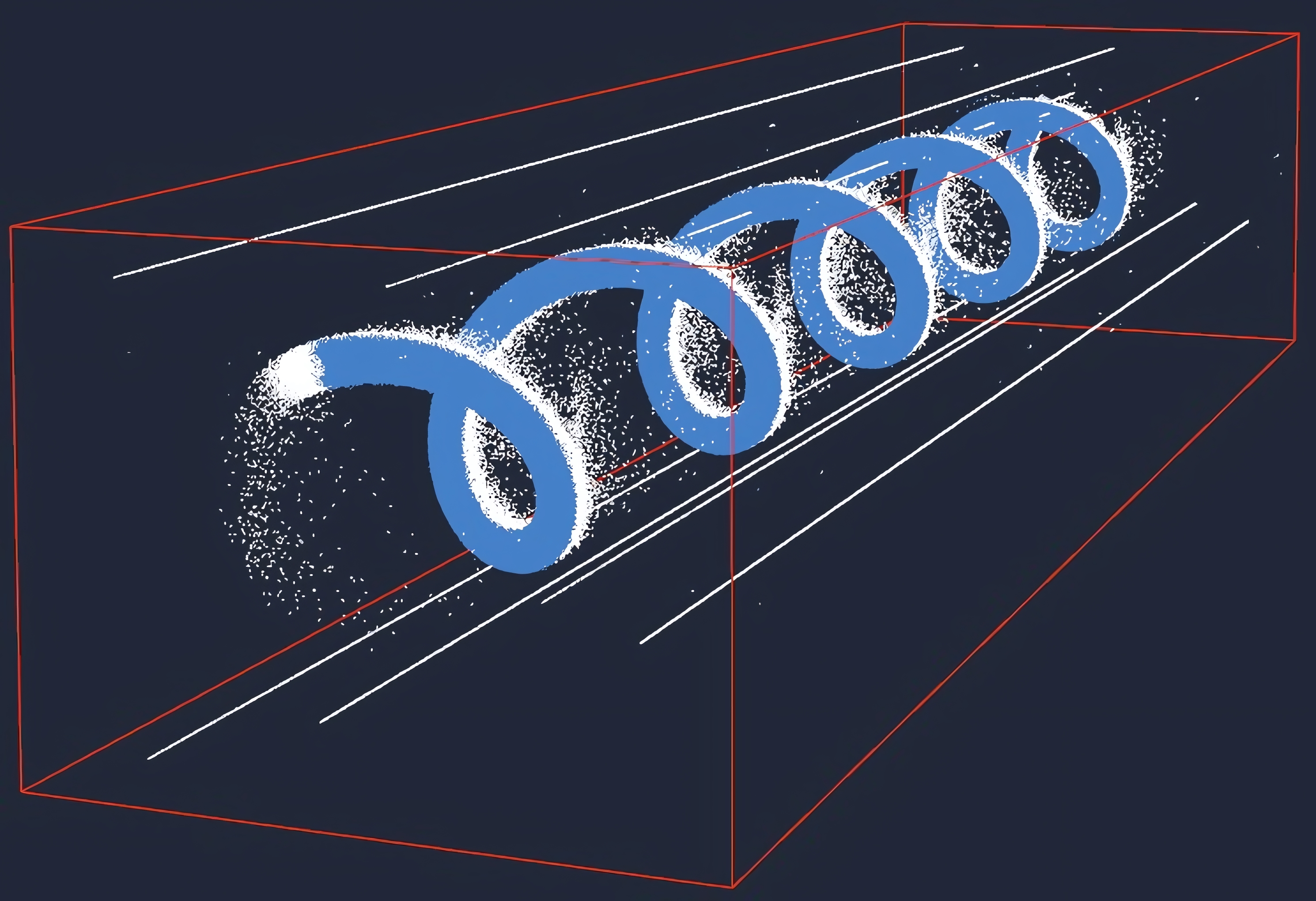}\\
\end{tabular}%
}\\
\hspace{3.5em} (a)~Target visualisation \hfill (b)~XYTime visualisation \hspace{4em}

\resizebox{\textwidth}{!}{%
\begin{tabular}{cc}
    \includegraphics[height=\textheight]{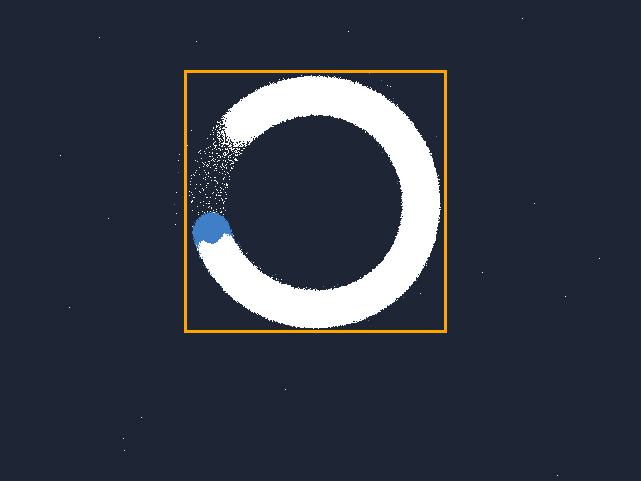} &
    \includegraphics[height=\textheight]{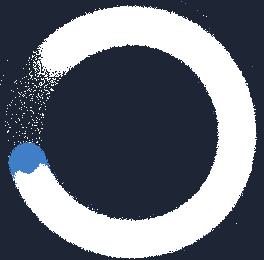}\\
\end{tabular}%
}\\
\hspace{5.2em} (c)~Aggregated events \hfill (d)~Region of Interest \hspace{3em}
\vspace{0.8em}\captionof{figure}{High-contrast dot (see Sec.~\ref{sec:dot}): (a)~target visualisation; (b)~events from a 250-millisecond window visualised in spatio-temporal space; (c)~aggregated events captured by the event camera ($640\times480$px) within a time window of length equal to $\frac{8}{10}$ of the period of the observed phenomenon, highlighted region of interest ($263\times263$px) shown in (d). Positive events are represented by white colour, and negative events are bright blue.}
\label{fig:dot_vis}
\end{table}
\FloatBarrier
\newpage

This experiment (see Sec.~\ref{sec:handspinner})  used event data from Prophesee's dataset, capturing a rotating three-blade hand fidget spinner, see Fig.~\ref{fig:handspinner_vis}.
\begin{table}[!htb]
\centering
\resizebox{\textwidth}{!}{%
\begin{tabular}{cc}
    \includegraphics[height=\textheight]{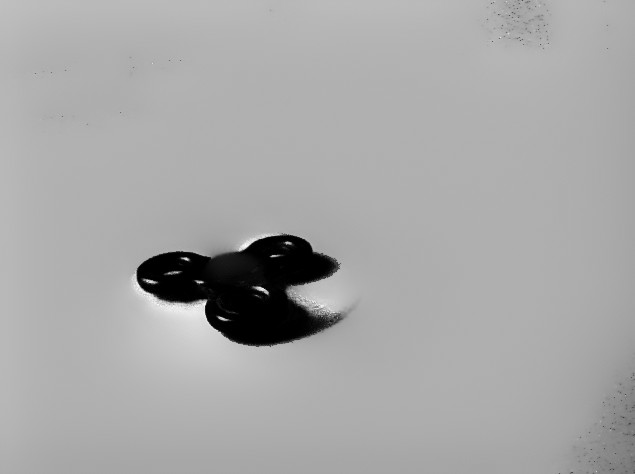} &
    \includegraphics[height=\textheight]{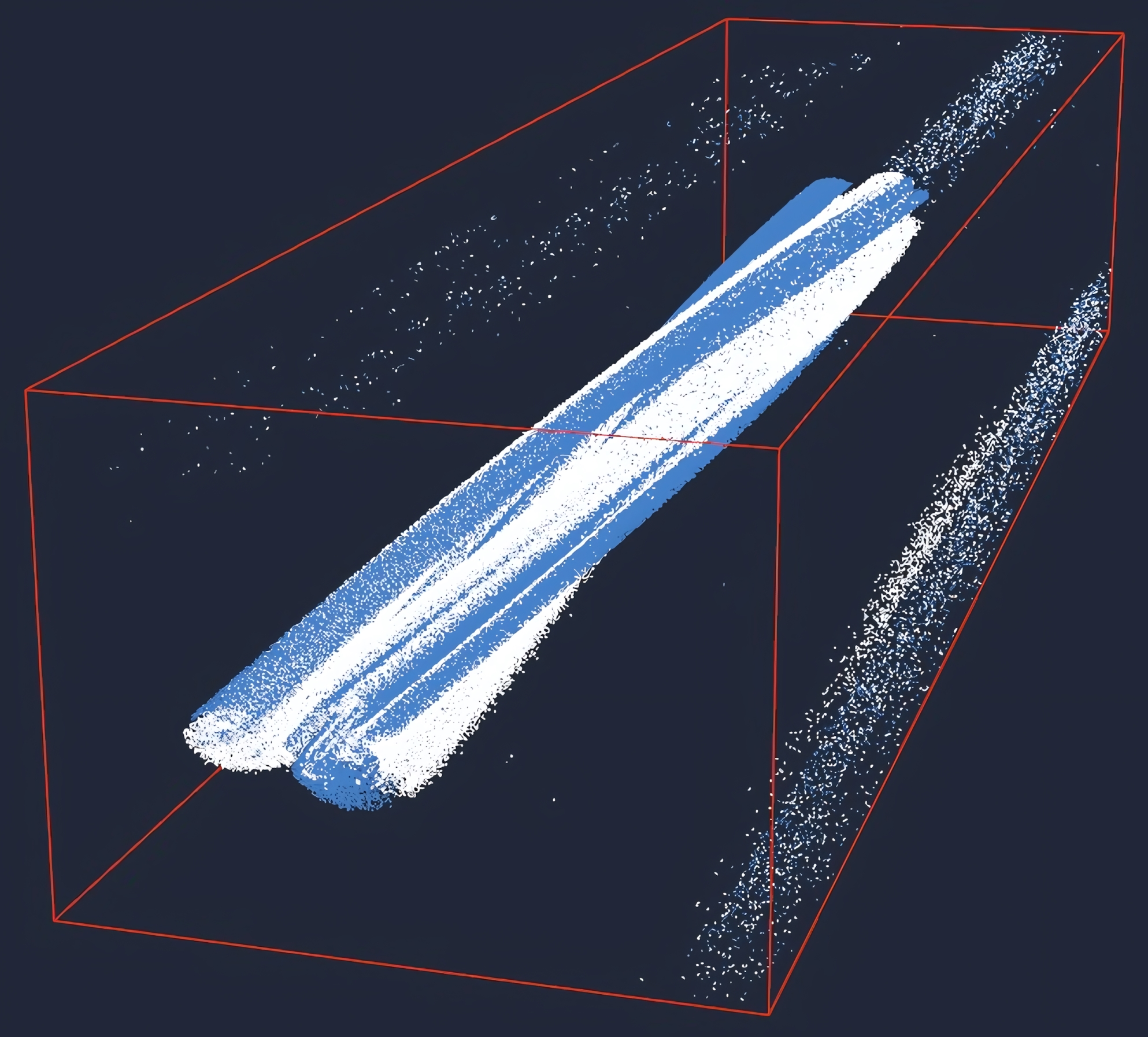}\\
\end{tabular}%
}\\
\hspace{5em} (a)~Target visualisation \hfill (b)~XYTime visualisation \hspace{2.5em}

\resizebox{\textwidth}{!}{%
\begin{tabular}{cc}
    \includegraphics[height=\textheight]{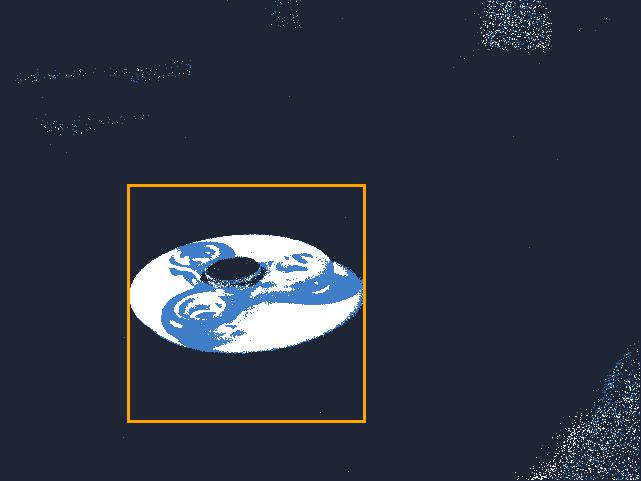} &
    \includegraphics[height=\textheight]{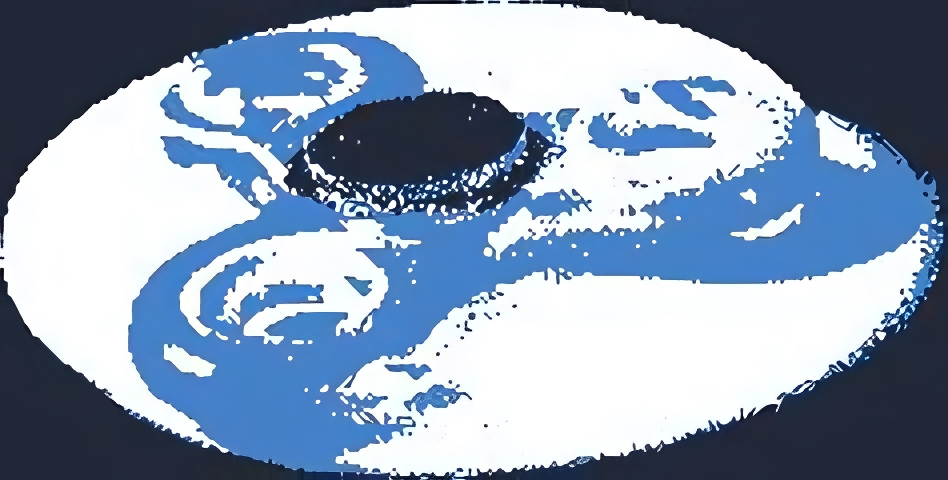}\\
\end{tabular}%
}\\
\hspace{2.5em} (c)~Aggregated events \hfill (d)~Region of Interest \hspace{6em}
\vspace{0.8em}\captionof{figure}{Hand fidget spinner (see Sec.~\ref{sec:handspinner}): (a)~target visualisation; (b)~events from a 250-millisecond window visualised in spatio-temporal space; (c)~aggregated events captured by the event camera ($640\times480$px) within a time window of length equal to the period of the observed phenomenon, highlighted region of interest ($239\times239$px) shown in (d). Positive events are represented by white colour, and negative events are bright blue.}
\label{fig:handspinner_vis}
\end{table}
\FloatBarrier
\newpage
The sequence used for this experiment captured the whirling behaviour for evading predators used by the Pholcus phalangioides spider (see Sec.~\ref{sec:spider}). 
A second rotating pattern is visible, caused by the shadow of the spider (see Fig.~\ref{fig:spider_vis}d,e).
\begin{table}[!htb]
\centering
\resizebox{\textwidth}{!}{%
\begin{tabular}{ccc}
    \includegraphics[height=\textheight]{images/spider_phys~1} & 
    \includegraphics[height=\textheight]{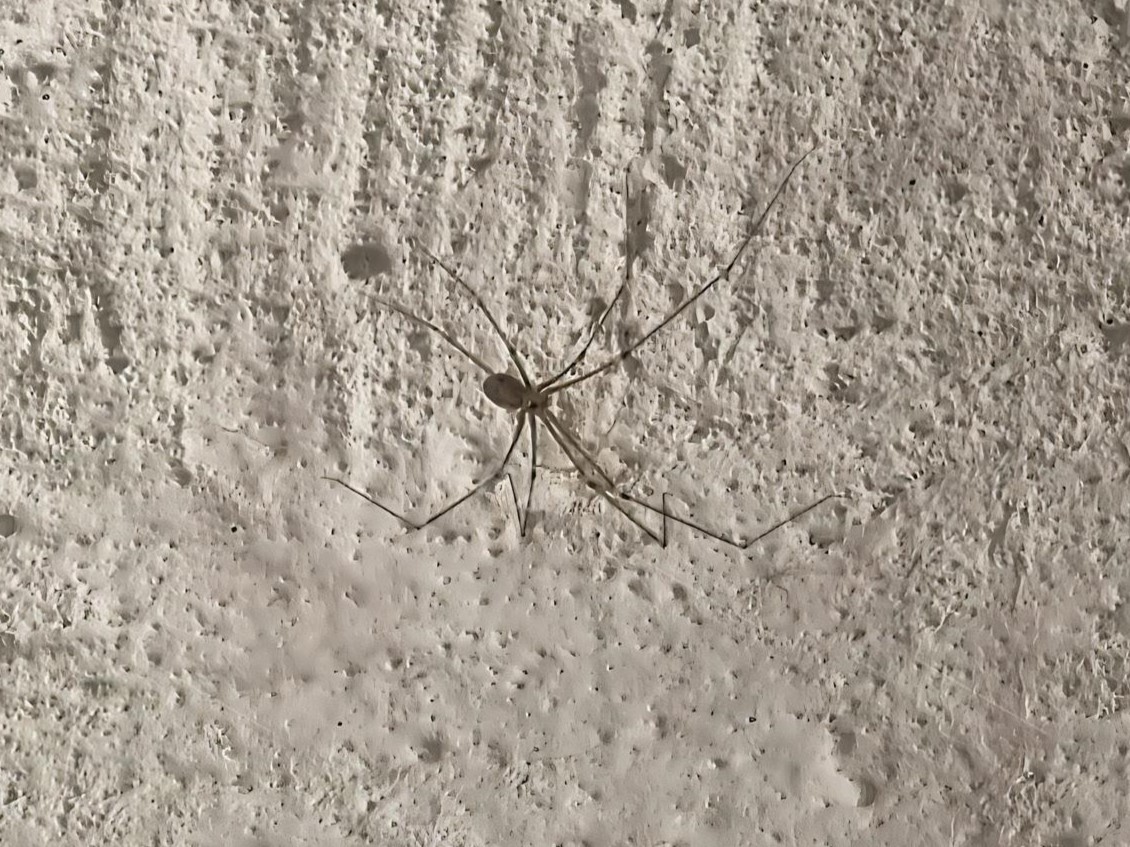}
    \includegraphics[height=\textheight]{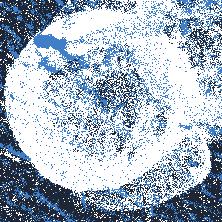}\\
\end{tabular}%
}\\
\hspace{2.5em} (a)~Physical setup \hspace{5em} (b)~Target close-up \hfill (c)~Region of Interest \hspace{0em}

\resizebox{\textwidth}{!}{%
\begin{tabular}{cc}
    \includegraphics[height=\textheight]{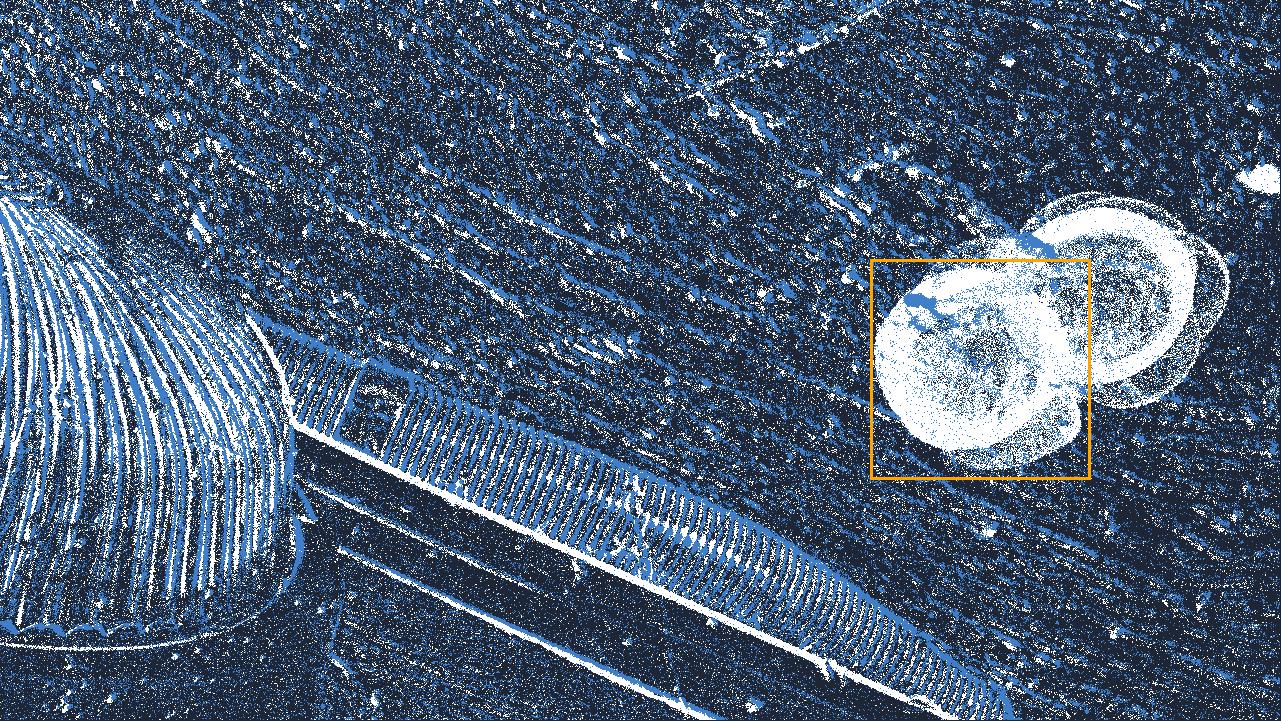} &
    \includegraphics[height=\textheight]{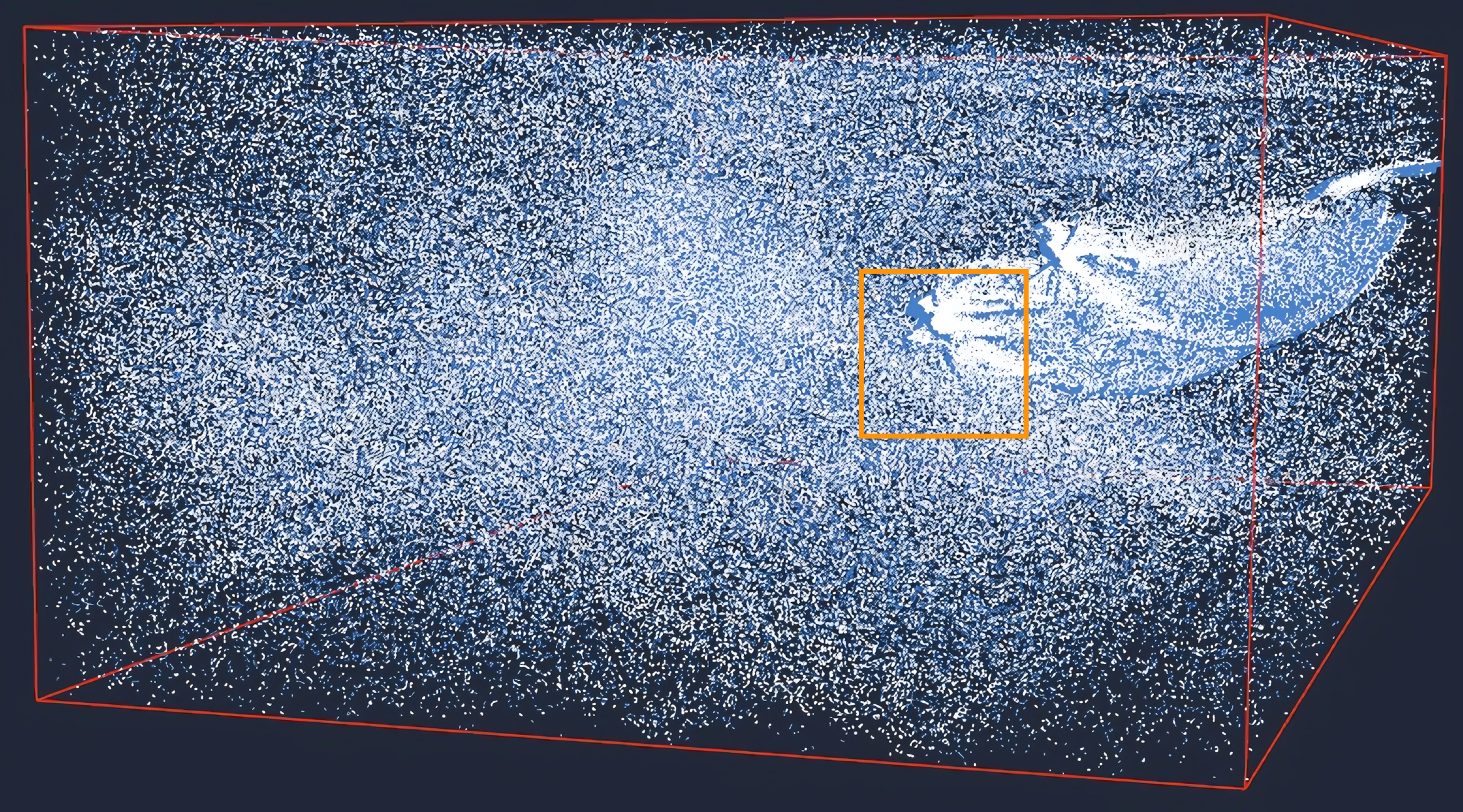}\\
\end{tabular}%
}\\
\hspace{4em} (d)~Aggregated events \hfill (e) XYTime visualisation \hspace{3.5em}
\vspace{0.8em}\captionof{figure}{Whirling Pholcus phalangioides (see Sec.~\ref{sec:spider}): (a)~physical setup; (b)~a close-up photo of the target; (d)~aggregated events captured by the event camera ($1280\times720$px) within a time window of length equal to the period of the observed phenomenon, highlighted region of interest ($221\times221$px) shown in (c); (e) events from a 250-millisecond window visualised in spatio-temporal space. Positive events are represented by white colour, and negative events are bright blue.}
\label{fig:spider_vis}
\end{table}
\FloatBarrier
\newpage
For the LED experiment (see Sec.~\ref{sec:led}), a circuit with a light-emitting diode (see Fig.~\ref{fig:led_vis}a) was used.
\begin{table}
\centering
\resizebox{\textwidth}{!}{%
\begin{tabular}{cc}
    \includegraphics[height=\textheight]{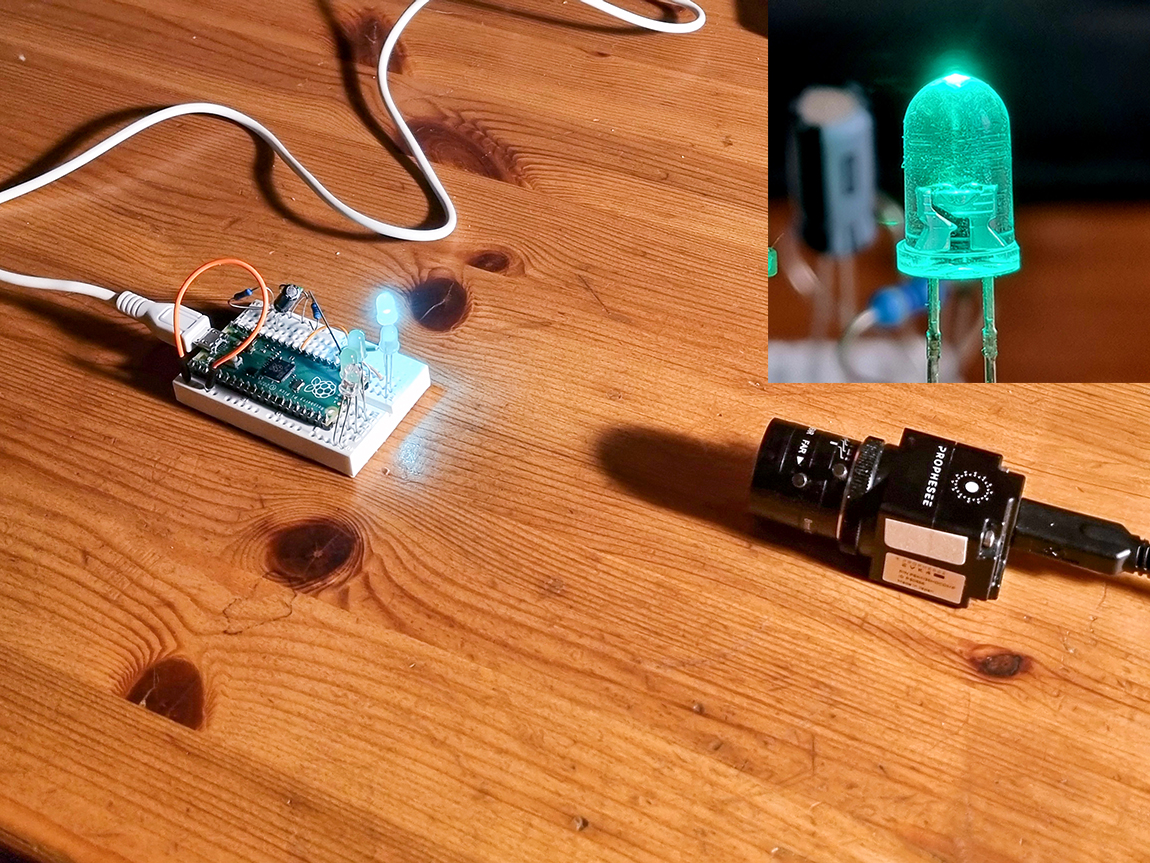} &
    \includegraphics[height=\textheight]{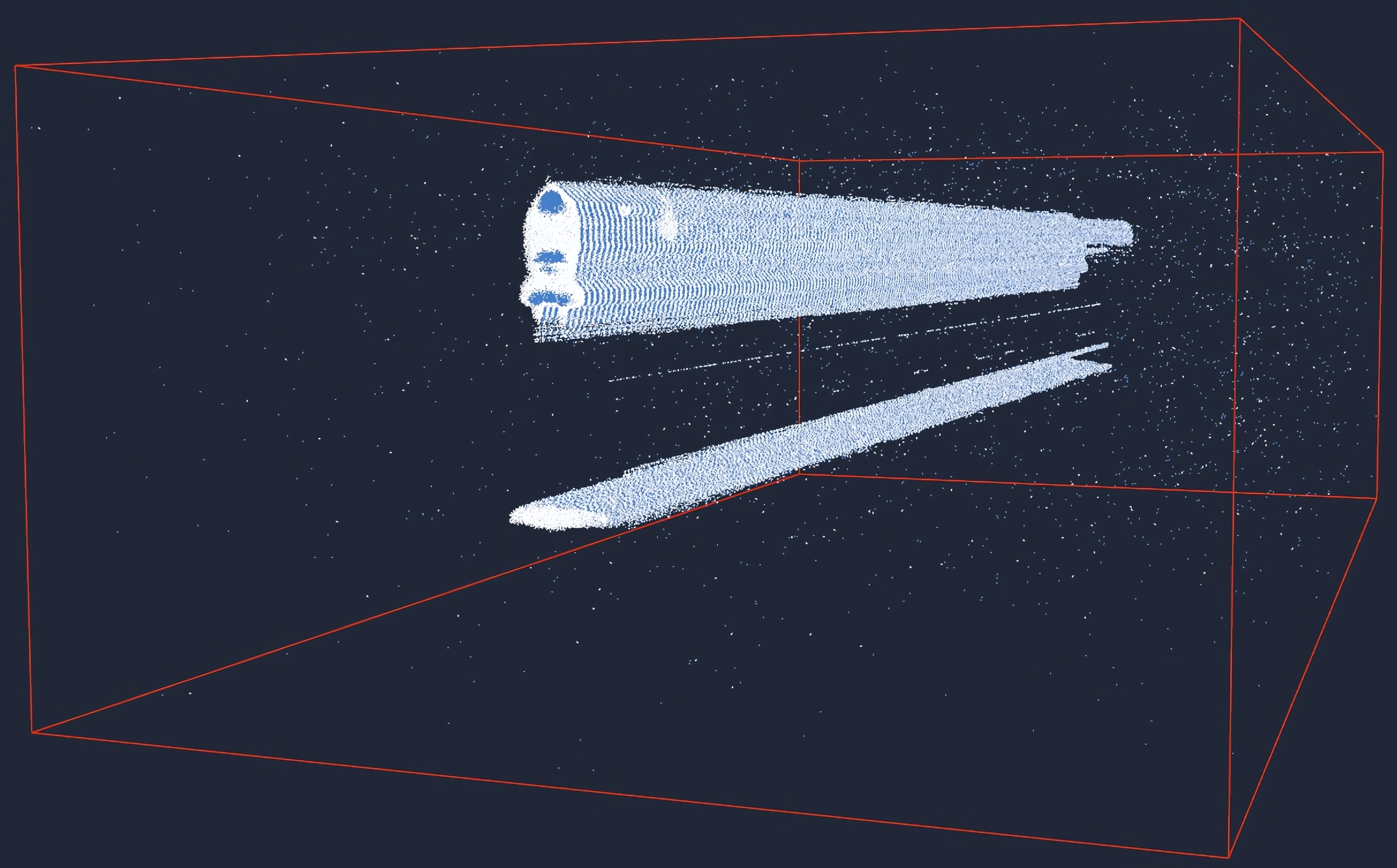}\\
\end{tabular}%
}\\
\hspace{4.5em} (a)~Physical setup \hfill (b)~XYTime visualisation \hspace{5em}

\resizebox{\textwidth}{!}{%
\begin{tabular}{cc}
\includegraphics[height=\textheight]{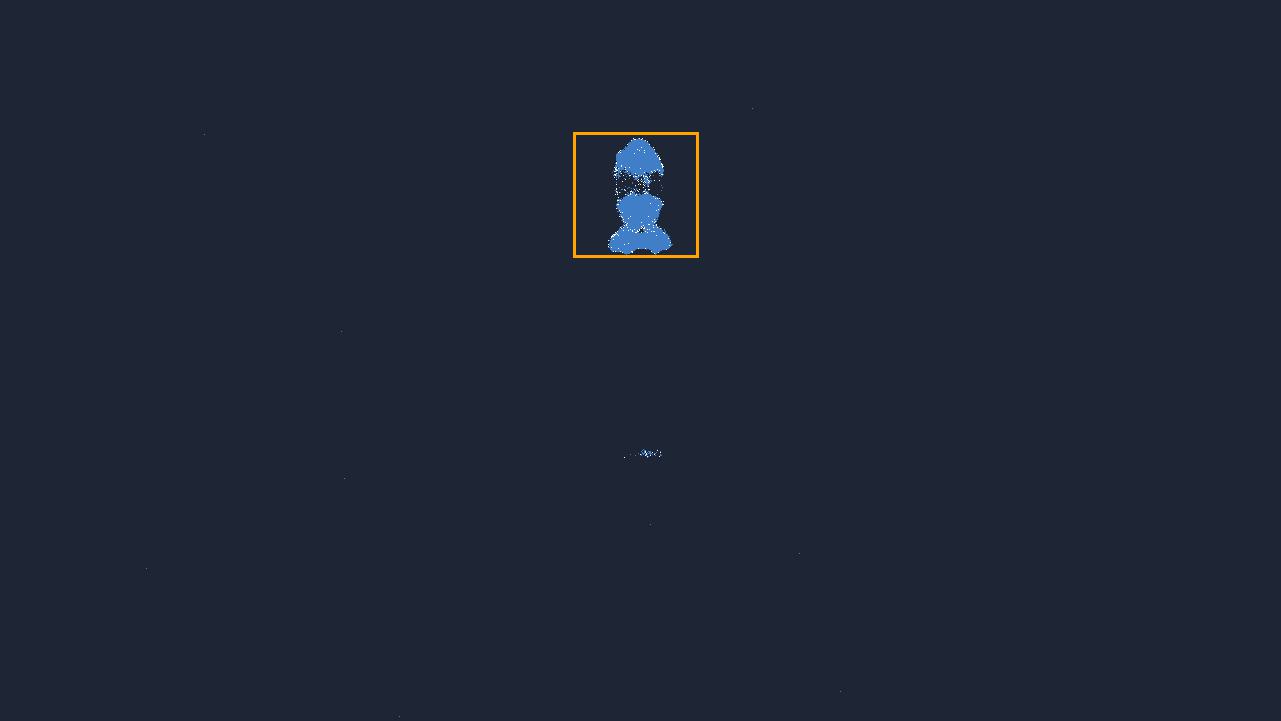} &
\includegraphics[height=\textheight]{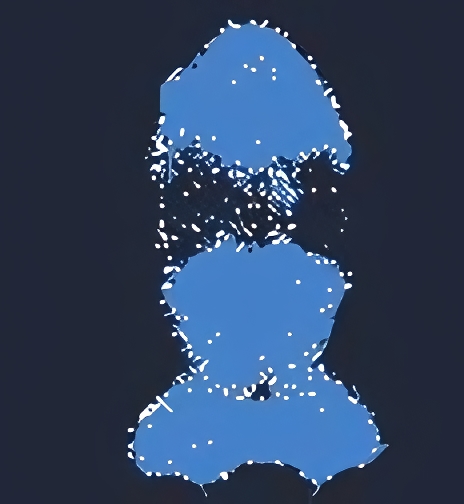}\\
\end{tabular}%
}\\
\hspace{7.5em} (c)~Aggregated events \hfill (d)~Region of Interest \hspace{1em}
\vspace{0.8em}\captionof{figure}{Flashing LED (see Sec.~\ref{sec:led}): (a)~physical setup and target detail; (b)~events from a 250-millisecond window visualised in spatio-temporal space; (c)~aggregated events captured by the event camera ($1280\times720$px) within a time window of length equal to the period of the observed phenomenon, highlighted region of interest ($126\times126$px) shown in (d). Positive events are represented by white colour, and negative events are bright blue.}
\label{fig:led_vis}
\end{table}
\FloatBarrier
\newpage
The sequence used for the 'Refreshing mobile phone screen' experiment (see Sec.~\ref{sec:screen}) captured the flicker of a mobile phone screen displaying a white colour at maximum brightness. A moving top-to-bottom pattern was visible when inspecting the event data (see Fig.~\ref{fig:screen_vis}b). 
\begin{table}[!htb]
\centering
\resizebox{\textwidth}{!}{%
\begin{tabular}{cc}
\includegraphics[height=\textheight]{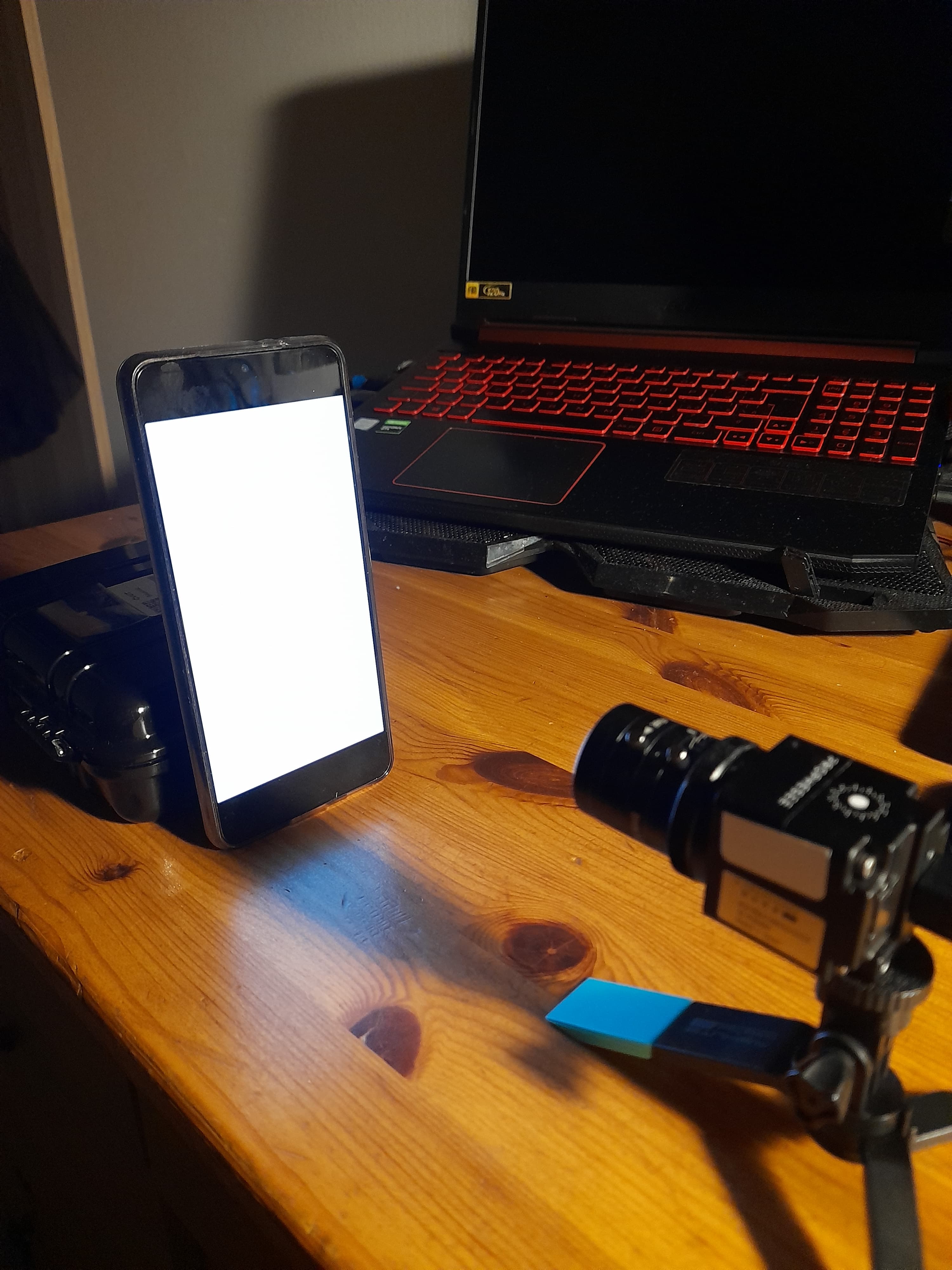} &
\includegraphics[height=\textheight]{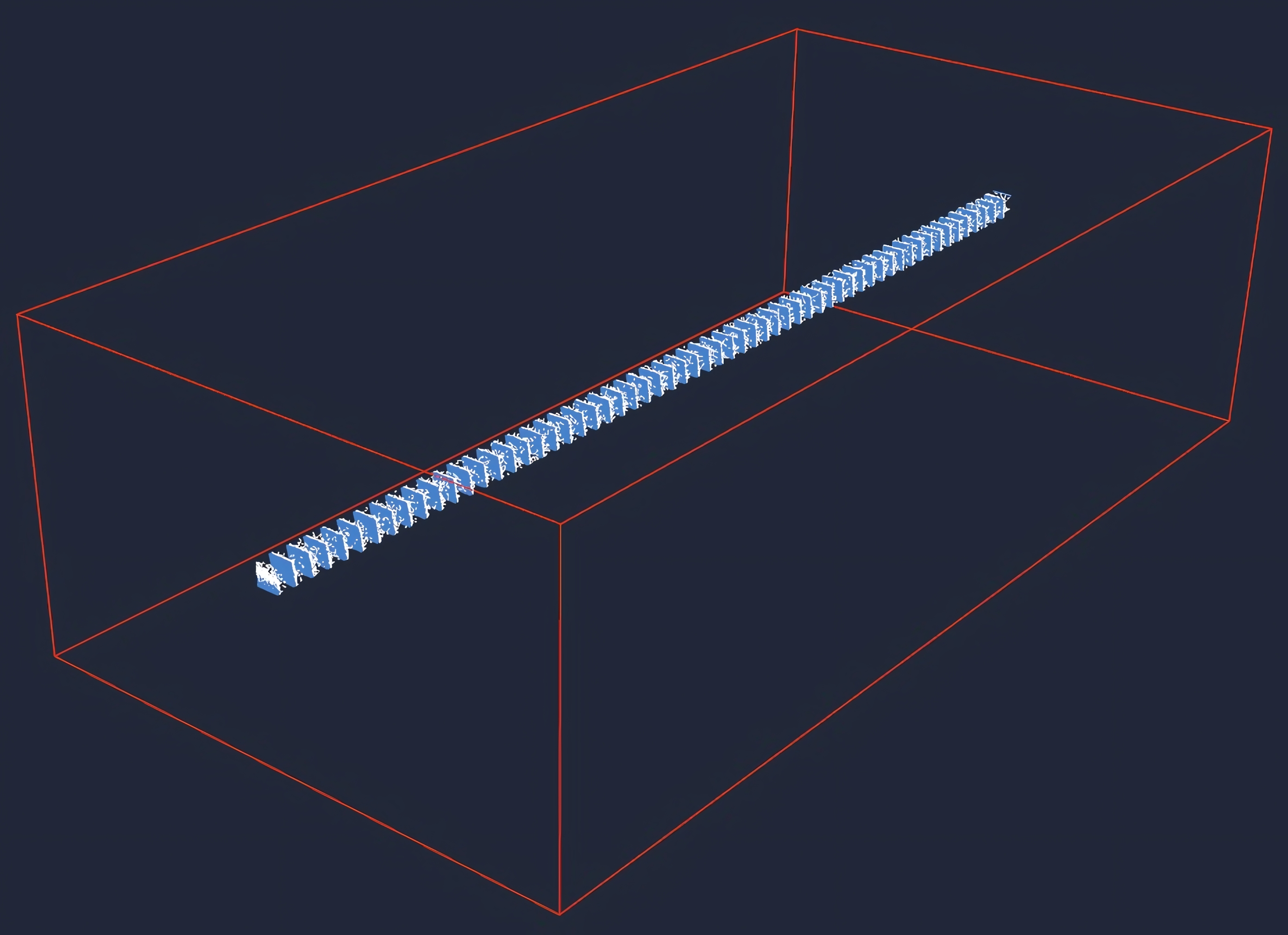}\\
\end{tabular}%
}\\
\hspace{2.5em} (a)~Physical setup \hfill (b)~XYTime visualisation \hspace{6.5em}

\resizebox{\textwidth}{!}{%
\begin{tabular}{cc}
\includegraphics[height=\textheight]{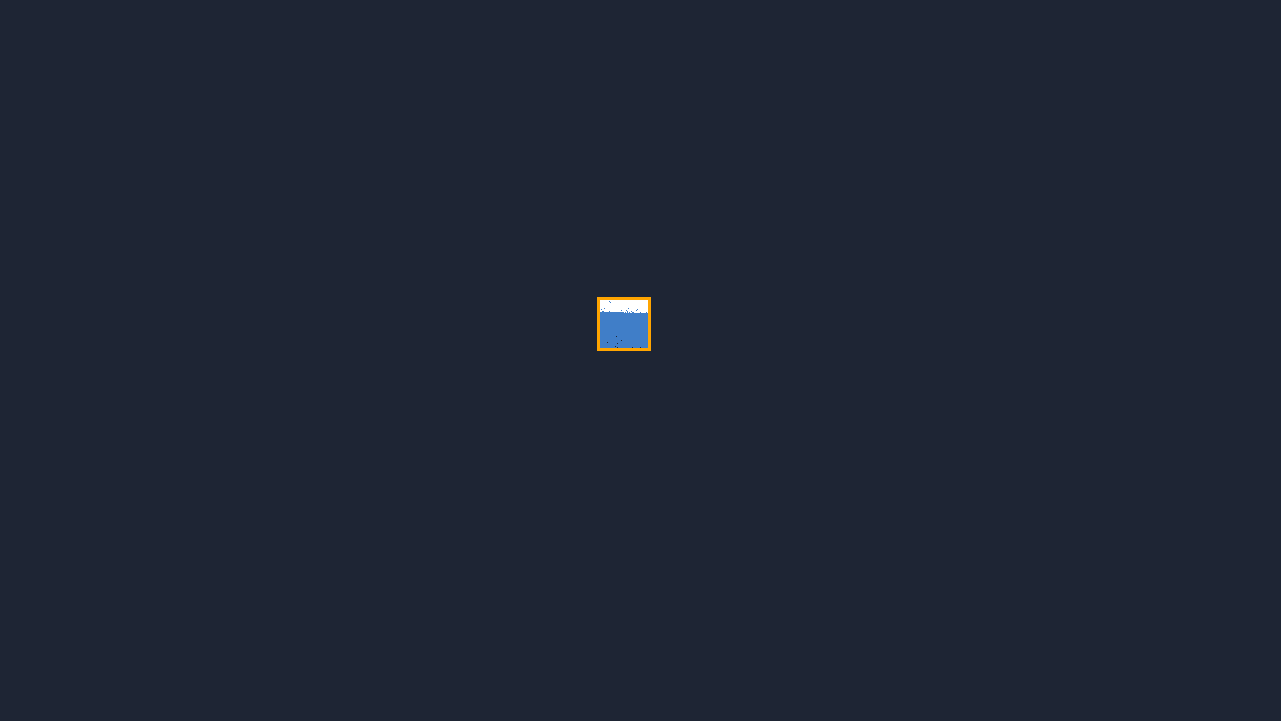} &
\includegraphics[height=\textheight]{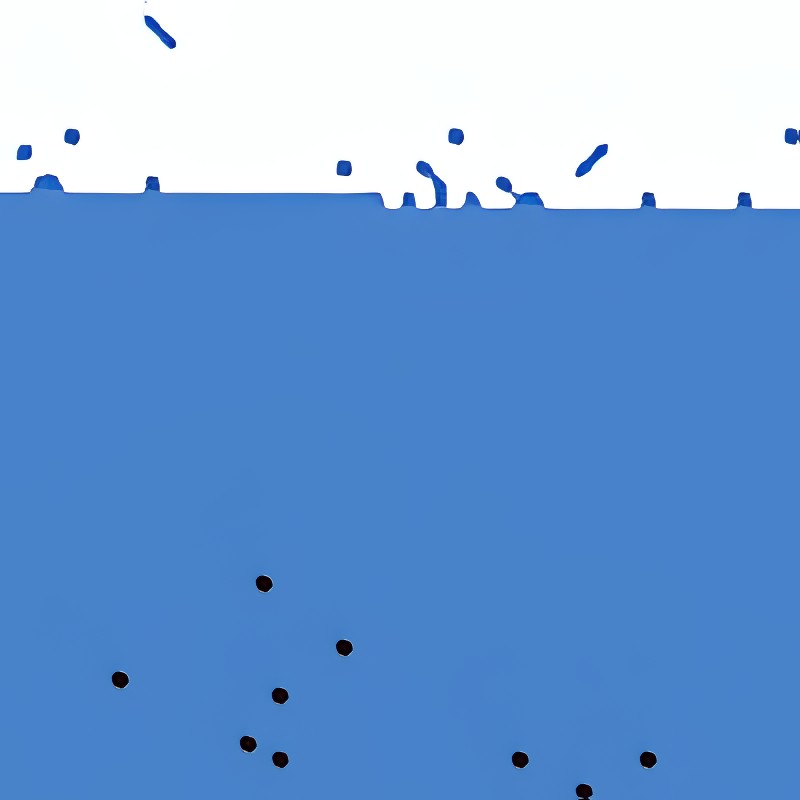}\\
\end{tabular}%
}\\
\hspace{7em} (c)~Aggregated events \hfill (d)~Region of Interest \hspace{1.3em}
\vspace{0.8em}\captionof{figure}{Refreshing mobile phone screen (see Sec.~\ref{sec:screen}): (a)~physical setup; (b)~events from a 250-millisecond window visualised in spatio-temporal space; (c)~aggregated events captured within hardware-level region of interest and a time window equal to $\frac{8}{10}$ of the period of the observed phenomenon, highlighted region of interest ($54\times54$px) shown in (d). Positive events are represented by white colour, and negative events are bright blue.}
\label{fig:screen_vis}
\end{table}
\FloatBarrier
\newpage

In the following experiment (see Sec.~\ref{sec:speaker}), we captured a vibrating diaphragm of a speaker with two large low-frequency drivers, see Fig.~\ref{fig:speaker_vis}a.

\begin{table}[!htb]
\centering
\resizebox{\textwidth}{!}{%
\begin{tabular}{cc}
\includegraphics[height=\textheight]{images/speaker_phys~1} &
\includegraphics[height=\textheight]{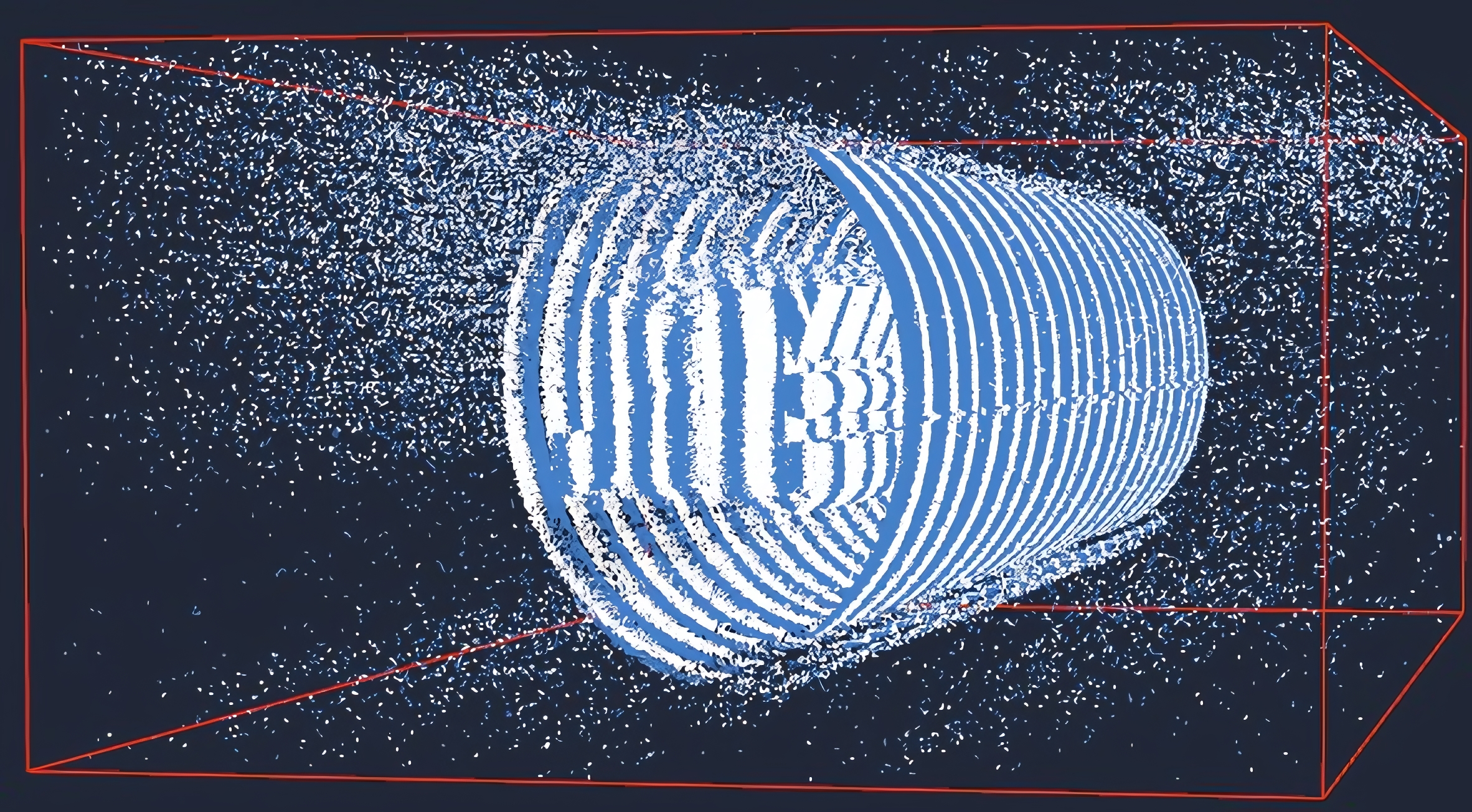}\\
\end{tabular}%
}\\
\hspace{4em} (a)~Physical setup \hfill (b)~XYTime visualisation \hspace{5em}

\resizebox{\textwidth}{!}{%
\begin{tabular}{cc}
\includegraphics[height=\textheight]{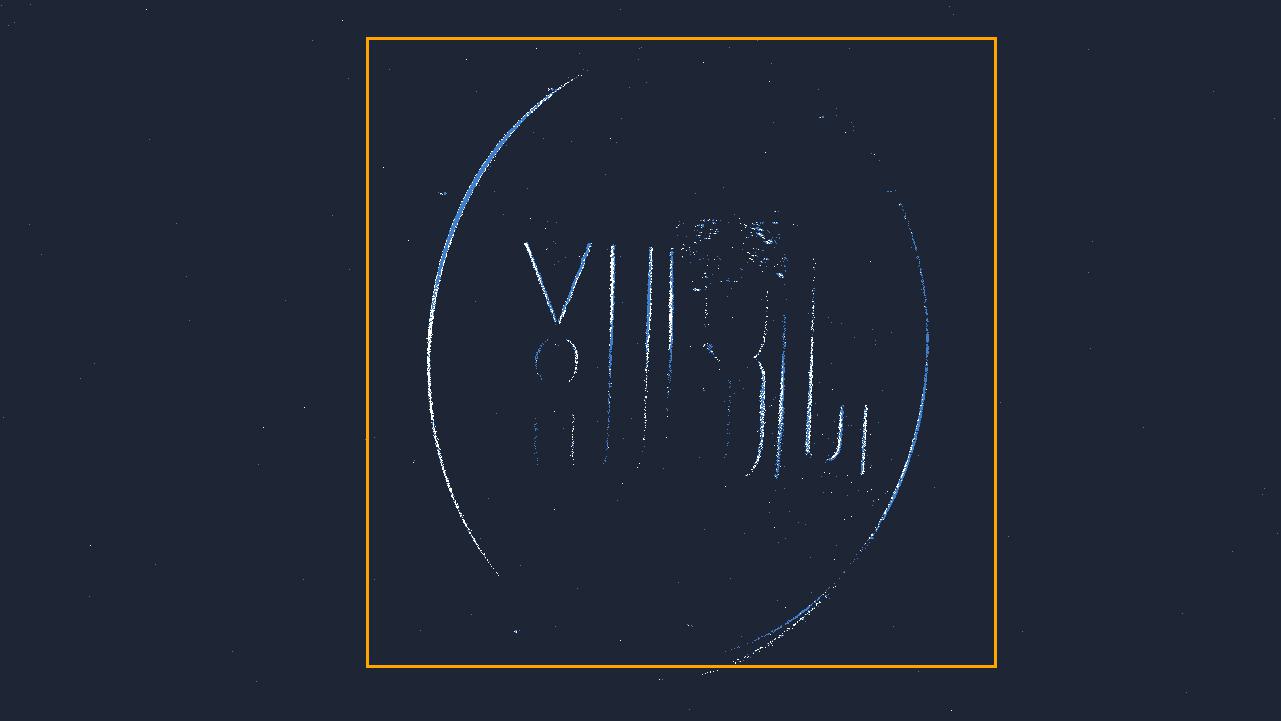} &
\includegraphics[height=\textheight]{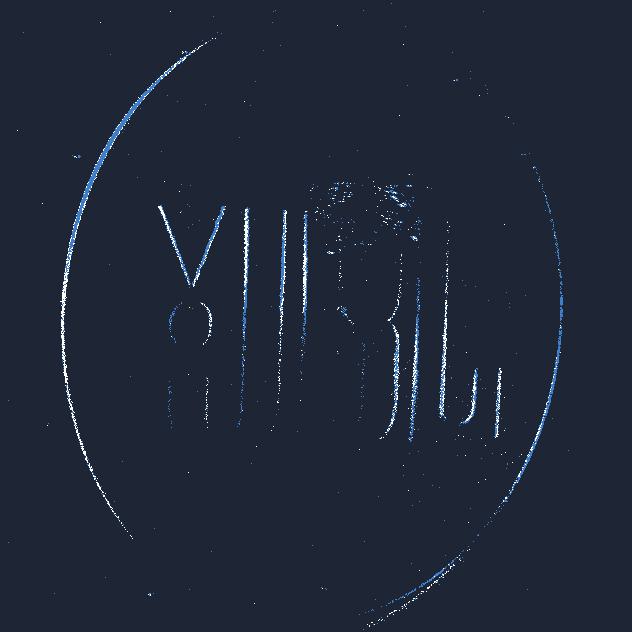}\\
\end{tabular}%
}\\
\hspace{7.2em} (c)~Aggregated events \hfill (d)~Region of Interest \hspace{2em}
\vspace{0.8em}\captionof{figure}{Speaker diaphragm (see Sec.~\ref{sec:speaker}): (a)~physical setup; (b)~events from a 250-millisecond window visualised in spatio-temporal space; (c)~aggregated events captured by the event camera ($1280\times720$px) within a time window of length equal to the period of the observed phenomenon, highlighted region of interest ($631\times631$px) shown in (d). Positive events are represented by white colour, and negative events are bright blue.}
\label{fig:speaker_vis}
\end{table}
\FloatBarrier
\newpage
For the 'Vibrating motor' experiment (see Sec.~\ref{sec:motor}) we used a sequence from Prophesee's dataset capturing a motor vibrating vertically at $40$~Hz; see Fig.~\ref{fig:motor_vis}.
\begin{table}[!htb]
\centering
\resizebox{\textwidth}{!}{%
\begin{tabular}{cc}
\includegraphics[height=\textheight]{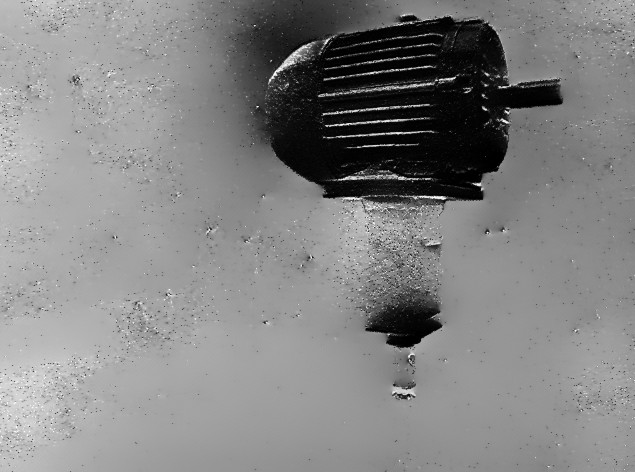} &
\includegraphics[height=\textheight]{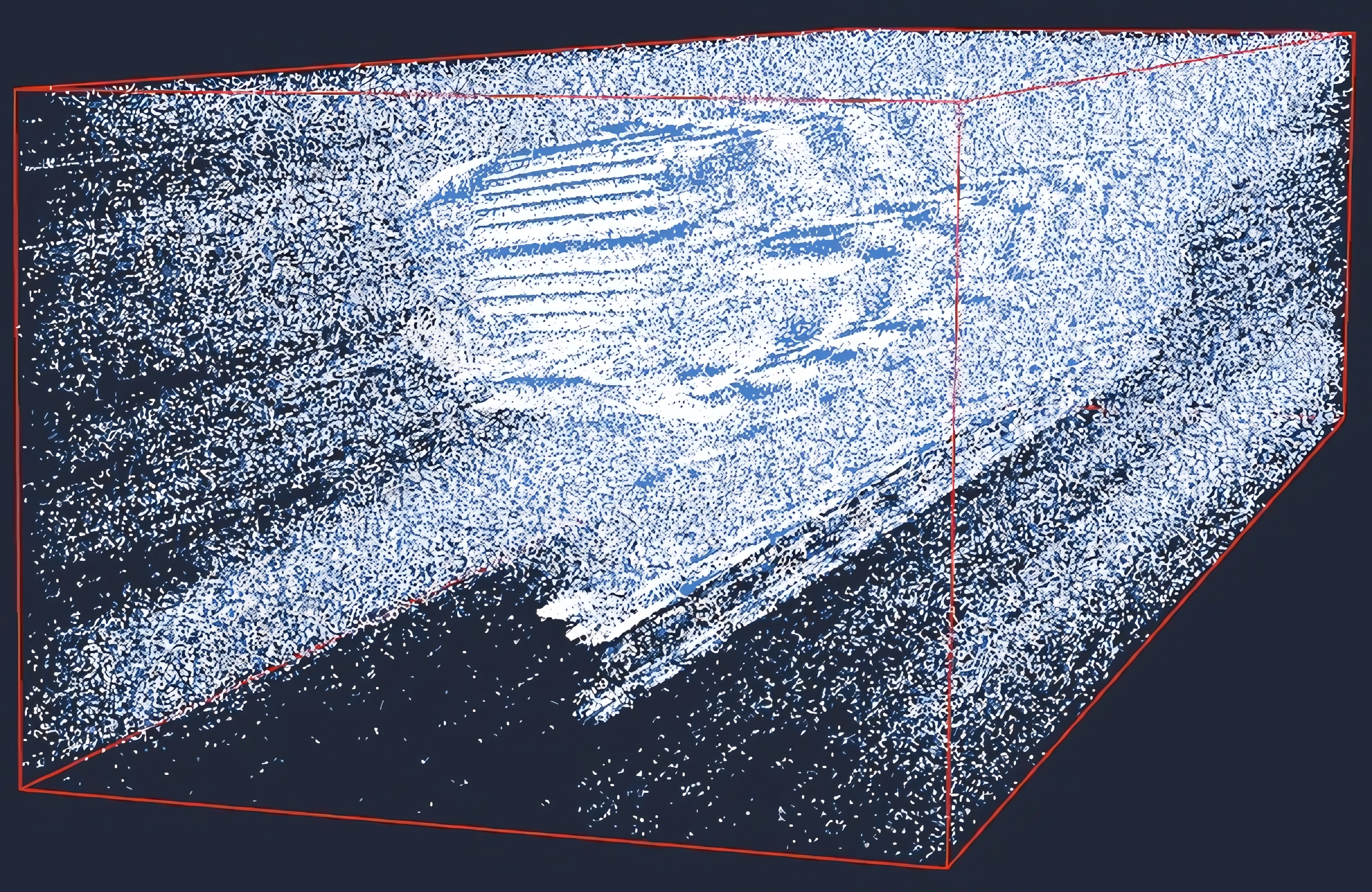}\\
\end{tabular}%
}\\
\hspace{3em} (a)~Target visualisation \hfill (b)~XYTime visualisation \hspace{4em}

\resizebox{\textwidth}{!}{%
\begin{tabular}{cc}
\includegraphics[height=\textheight]{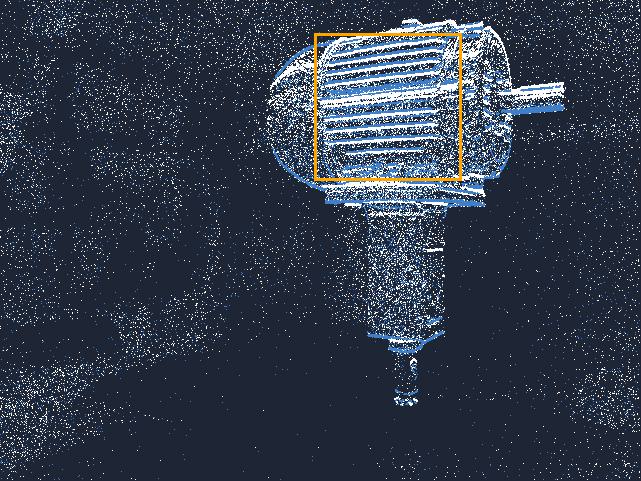} &
\includegraphics[height=\textheight]{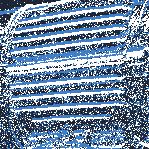}\\
\end{tabular}%
}\\
\hspace{5em} (c)~Aggregated events \hfill (d)~Region of Interest \hspace{2.5em}
\vspace{0.8em}\captionof{figure}{Vibrating motor (see Sec.~\ref{sec:motor}): (a)~target visualisation; (b)~events from a 250-millisecond window visualised in spatio-temporal space; (c)~aggregated events captured by the event camera ($640\times480$px) within a time window of length equal to the period of the observed phenomenon, highlighted region of interest ($148\times148$px) shown in (d). Positive events are represented by white colour, and negative events are bright blue.}
\label{fig:motor_vis}
\end{table}
\FloatBarrier
\newpage
The sequence used for this experiment (see Sec.~\ref{sec:chain_side}) captured a bicycle chain from a close-up side view (see Fig.~\ref{fig:chain_side_vis}a). 
\begin{table}[!htb]
\centering
\resizebox{\textwidth}{!}{%
\begin{tabular}{cc}
\includegraphics[height=\textheight]{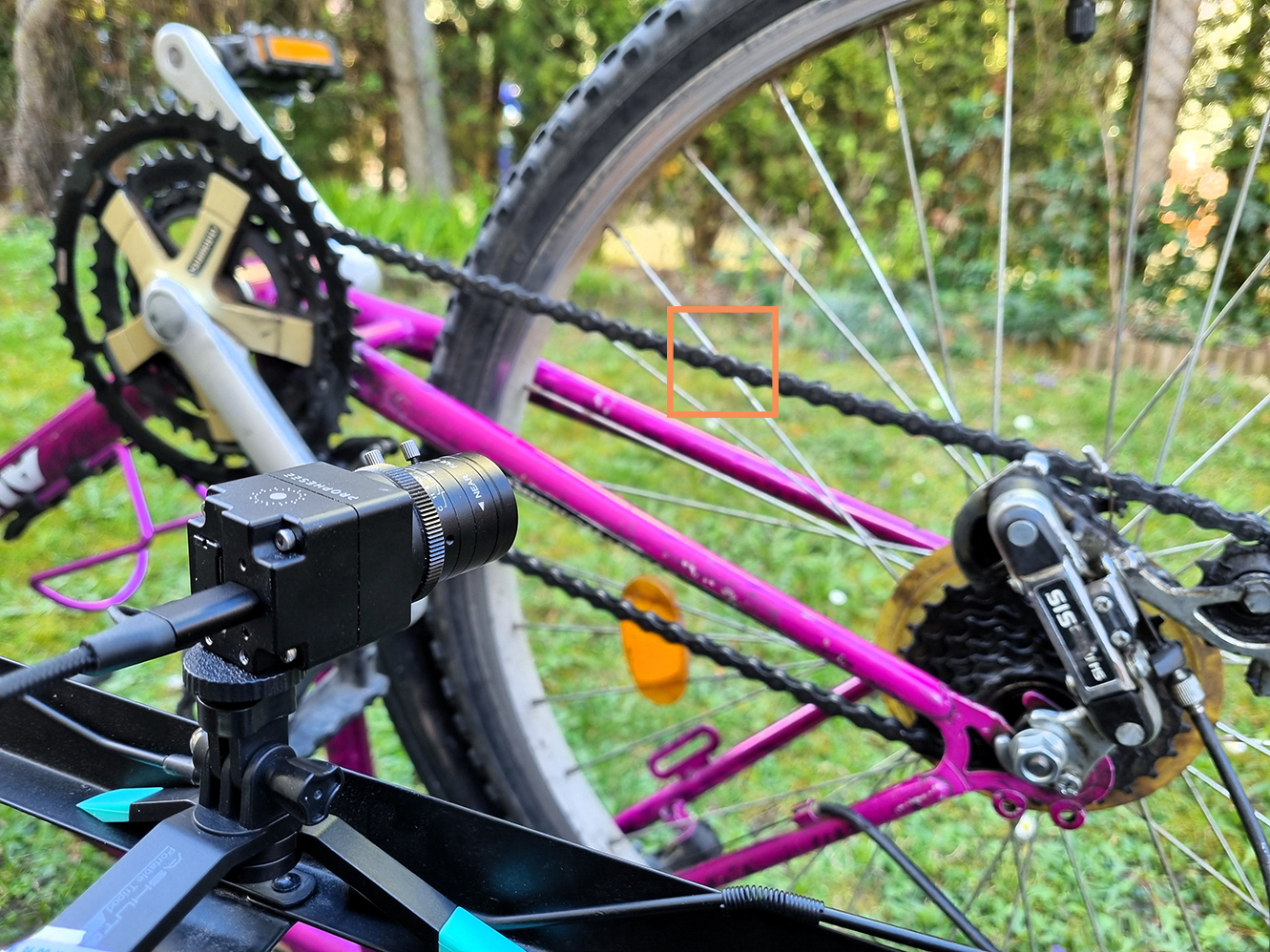} &
\includegraphics[height=\textheight]{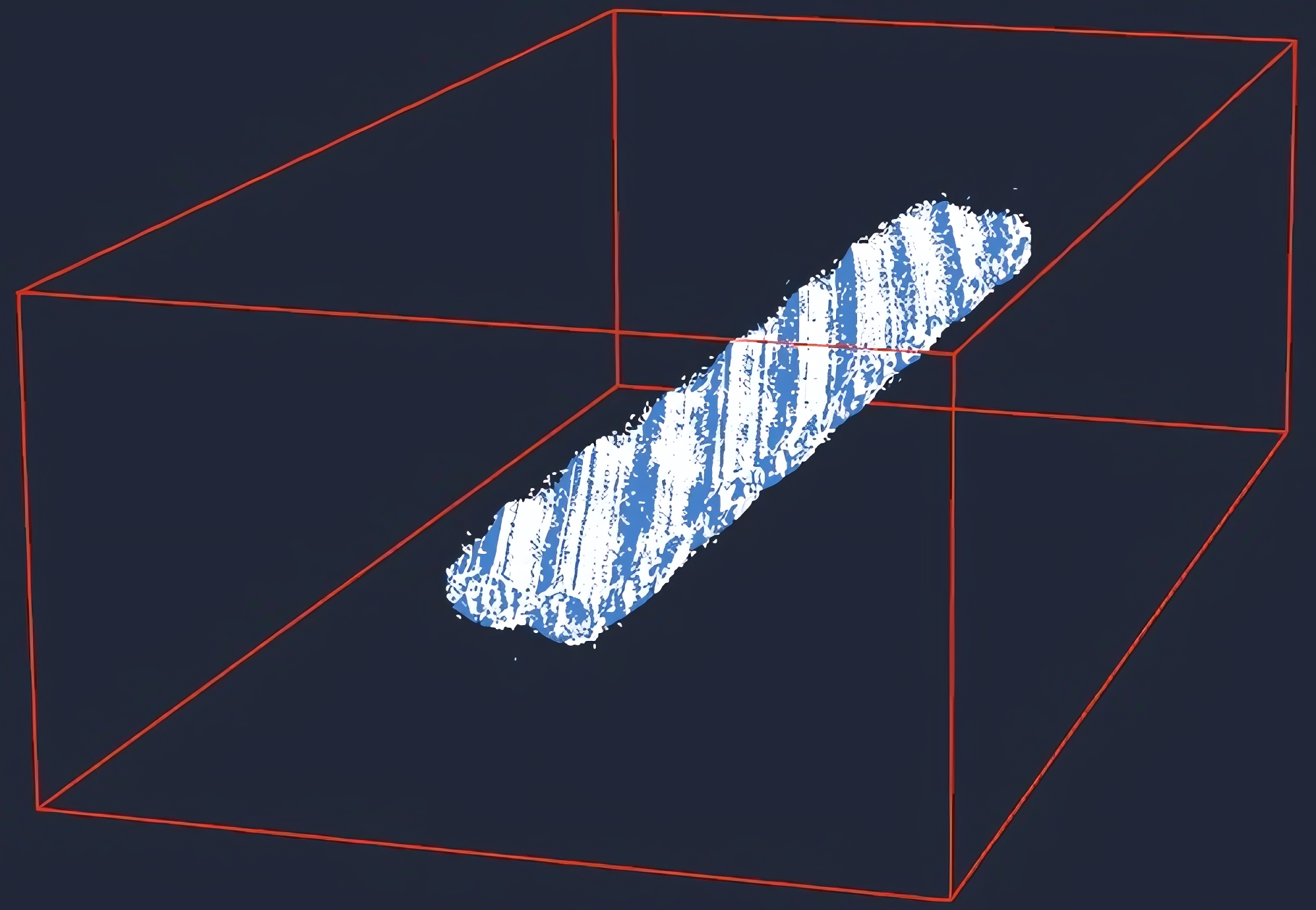}\\
\end{tabular}%
}\\
\hspace{5em} (a)~Physical setup \hfill (b)~XYTime visualisation \hspace{4em}

\resizebox{\textwidth}{!}{%
\begin{tabular}{cc}
\includegraphics[height=\textheight]{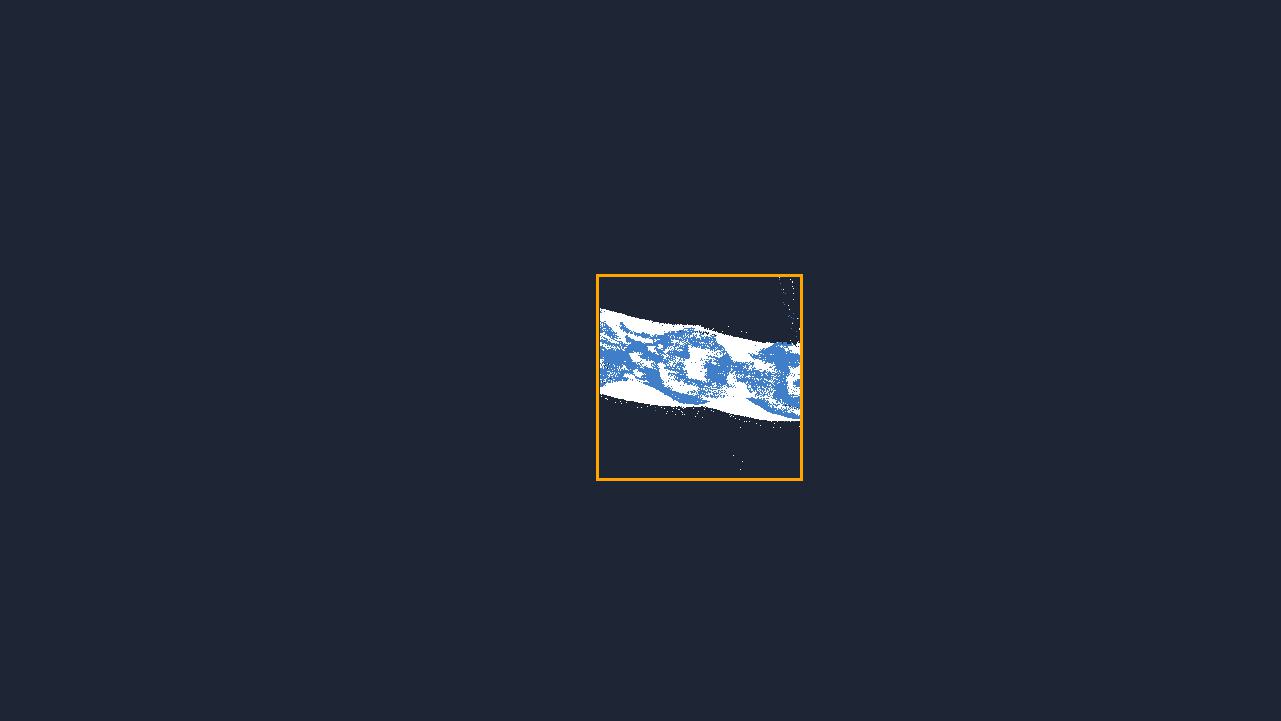} &
\includegraphics[height=\textheight]{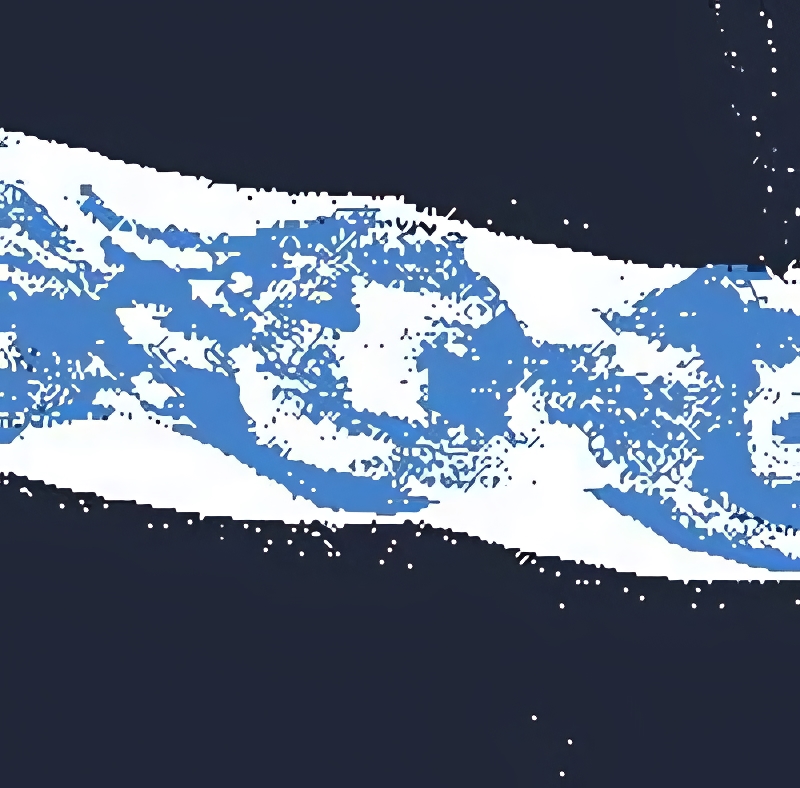}\\
\end{tabular}%
}\\
\hspace{7em} (c)~Aggregated events \hfill (d)~Region of Interest \hspace{1.5em}
\vspace{0.8em}\captionof{figure}{Bike chain from side view (see Sec.~\ref{sec:chain_side}): (a)~physical setup; (b)~events from a 250-millisecond window visualised in spatio-temporal space; (c)~aggregated events captured within a hardware-level region of interest and a time window equal to the period of the observed phenomenon, highlighted region of interest ($204\times204$px) shown in (d). Positive events are represented by white colour, and negative events are bright blue.}
\label{fig:chain_side_vis}
\end{table}
\FloatBarrier
\newpage
The sequence for the final experiment (see Sec.~\ref{sec:chain_top}) captured the bike chain from a top view (see Fig.~\ref{fig:chain_top_vis}a).
\begin{table}[!htb]
\centering
\resizebox{\textwidth}{!}{%
\begin{tabular}{cc}
\includegraphics[height=\textheight]{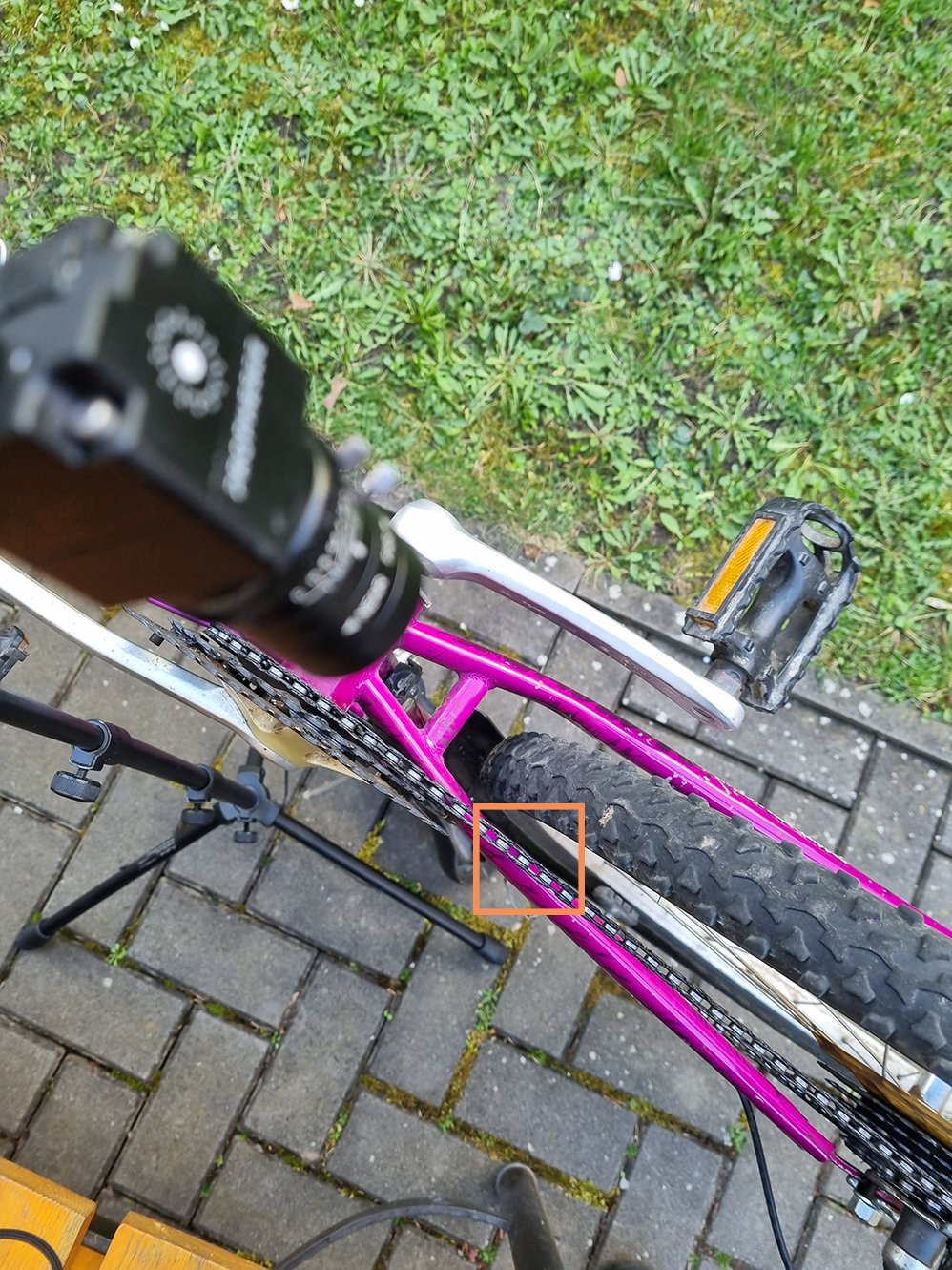} &
\includegraphics[height=\textheight]{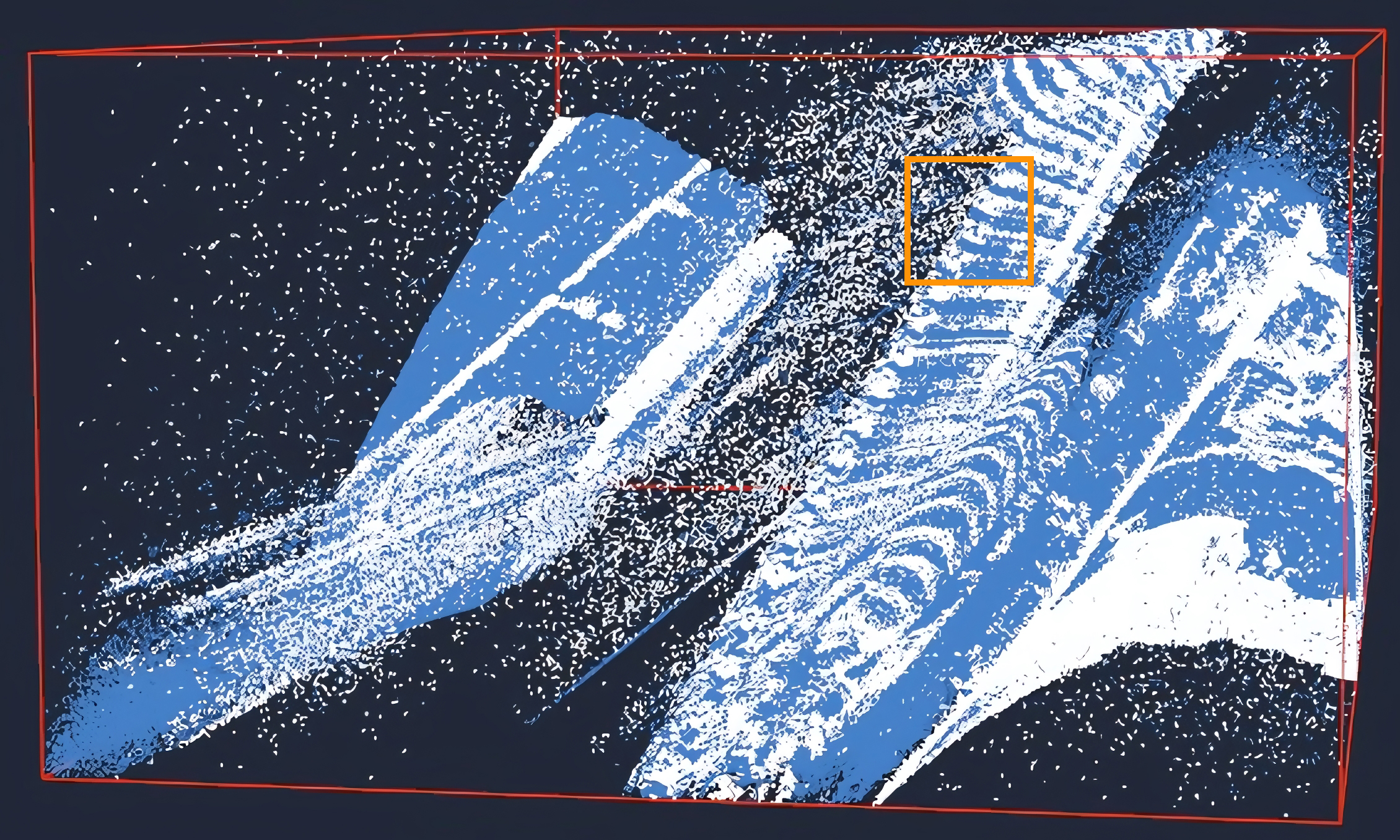}\\
\end{tabular}%
}\\
\hspace{1.5em} (a)~Physical setup \hfill (b)~XYTime visualisation \hspace{7.5em}

\resizebox{\textwidth}{!}{%
\begin{tabular}{cc}
\includegraphics[height=\textheight]{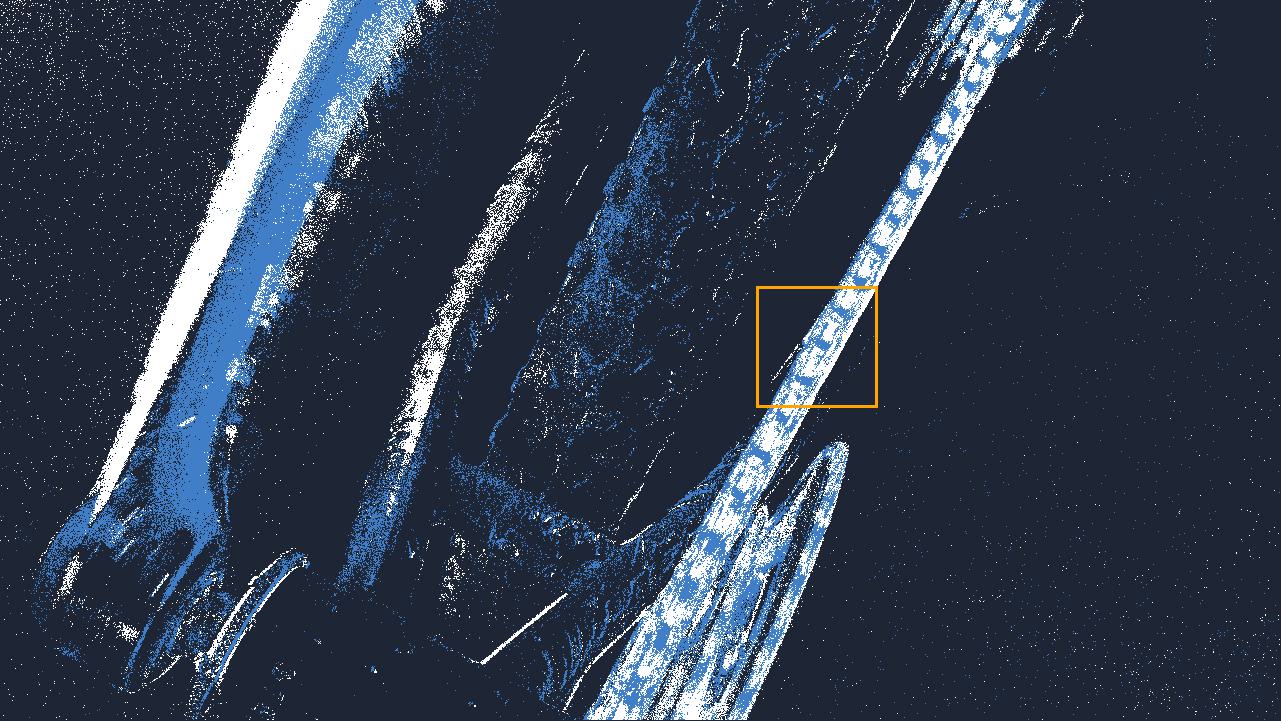} &
\includegraphics[height=\textheight]{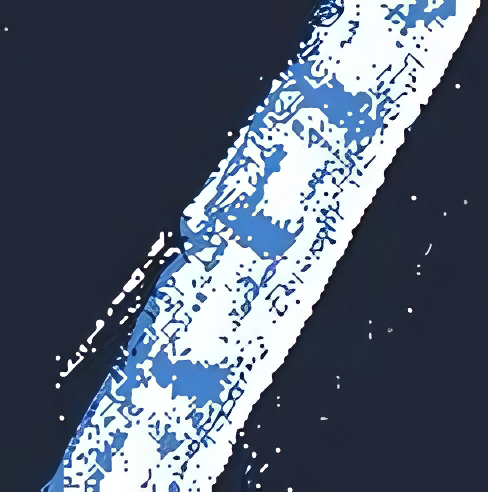}\\
\end{tabular}%
}\\
\hspace{7.5em} (c)~Aggregated events \hfill (d)~Region of Interest \hspace{1.3em}
\vspace{0.8em}\captionof{figure}{Bike chain from a top view (see Sec.~\ref{sec:chain_top}): (a)~physical setup; (b)~events from a 250-millisecond window visualised in spatio-temporal space; (c)~aggregated events captured by the event camera ($1280\times720$px) within a time window of lenght equal to the period of the observed phenomenon, highlighted region of interest ($180\times180$px) shown in (d). Positive events are represented by white colour, and negative events are bright blue.}
\label{fig:chain_top_vis}
\end{table}